%% file: main.tex
\documentclass{applemlr}
\pdfmapfile{+cm-super-t1.map}
\pdfmapfile{+cm-super-ts1.map}

\usepackage{amsmath,amsfonts}
\usepackage{mathtools}

\usepackage{hyperref}
\usepackage{url}
\usepackage{booktabs}
\usepackage{longtable}
\usepackage{array}
\usepackage{multirow}
\usepackage{graphicx}
\usepackage{subcaption}
\usepackage{xcolor}
\usepackage{algorithm}
\usepackage{algorithmic}

\title{Selection-Based Structured Reasoning: \\ Toward Efficient Multimodal Search Agents}

\author[1,2]{Feiyu Gavin Zhu}
\author[1\ddagger*]{Xiaoyu Zhu}
\author[1]{Jiqi Yang}
\author[1]{Rui Yang}
\author[1]{Arnab Kumar Mondal}
\author[1]{Yancheng Wang}
\author[1]{Xinke Deng}
\author[2]{Jean Oh}
\author[2]{Reid Simmons}
\author[1]{Joerg Liebelt}
\author[1]{Xiang Kong}
\author[1*]{Zhongyu Jiang}
\affiliation[1]{Apple}
\affiliation[2]{Carnegie Mellon University}
\contribution[\ddagger]{Project Lead}
\contribution[*]{Corresponding Author}

\abstract{\begingroup
  \renewenvironment{abstract}{}{}
  \input{section/0_abstract}
\endgroup}
\setcitestyle{authoryear,round,citesep={;},aysep={,},yysep={;}}

\definecolor{linkblue}{RGB}{0,102,204}
\hypersetup{
  pdfauthor={Gavin Zhu; Xiaoyu Zhu; Jiqi Yang; Rui Yang; Arnab Kumar Mondal; Yancheng Wang; Xinke Deng; Jean Oh; Reid Simmons; Joerg Liebelt; Xiang Kong; Zhongyu Jiang},
  pdftitle={Selection-Based Structured Reasoning: Toward Efficient Multimodal Search Agents},
  colorlinks=true,
  citecolor=linkblue,
  linkcolor=black,
  urlcolor=black,
  filecolor=black
}
\patchcmd{\mymaketitle}{\vskip 0.5cm}{}{}{\errmessage{Title spacing patch failed}}
\patchcmd{\mymaketitle}{\vspace*{0.65cm}}{}{}{\errmessage{Metadata spacing patch failed}}
\begin{document}

\maketitle
\input{section/1_intro}

\input{section/2_related}
\input{section/3_method}
\input{section/4_exp}
\input{section/5_conclusion}

\input{section/6_statements}

\bibliography{references}
\bibliographystyle{references}

\clearpage
\appendix
\section*{Appendix}
\input{appendix/9_inference_algorithm}
\clearpage 
\input{appendix/1_selector_gradient}

\input{appendix/6_inference_cost}
\input{appendix/2_search_benchmarks}
\input{appendix/4_strategy_library}
\input{appendix/0_training_details}
\input{appendix/3_reasoning_comparison}
\begingroup
\renewcommand\thefootnote{}
\makeatletter
\renewcommand\@makefntext[1]{\noindent#1}
\makeatother
\footnotetext{\textcolor[HTML]{6E6E73}{\sffamily Apple and the Apple logo are trademarks of Apple Inc., registered in the U.S. and other countries and regions.}}
\endgroup
\end{document}

%% file: section/0_abstract.tex
\begin{abstract}

Multimodal agents commonly generate freeform reasoning before each action.
For small models, limited model capacity can result in lengthy reasoning that provides little useful guidance for action generation while incurring substantial inference cost.
To address this challenge, we introduce Selection-based Structured Reasoning (SSR), a framework that reformulates reasoning as selection instead of open-ended generation.
SSR represents recurring high-level reasoning as pre-specified, reusable natural-language candidates.
At each turn, the model selects from the reasoning candidates based on their likelihoods given the current context, without an auxiliary task head.
Using pre-specified reasoning traces enables parallel scoring, where teacher-forced prefilling computes token likelihoods concurrently within and across candidates using a shared context KV cache.
We evaluated SSR on seven multimodal search benchmarks using 2B and 4B models.
Across multiple reinforcement learning objectives and supervised fine-tuning, SSR delivers significant efficiency gains without sacrificing task performance.
SSR achieves an average success rate competitive with leading search agents of the same scale, while reducing per-turn reasoning latency by over 90\% and total per-question model inference latency by 28–54\%.
Project page:  \url{https://zfy0314.github.io/ssr-webpage/}

\end{abstract}

%% file: section/1_intro.tex
\section{Introduction}

Multimodal agents turn textual and visual observations into actions: they analyze multimodal inputs, gather information, and use tools to resolve tasks that require multi-turn interactions~\citep{he2024webvoyager,chng2025sensenova}.
A common agent design interleaves observations, freeform reasoning, and actions~\citep{yao2022react}.
The reasoning step lets the model assess the available evidence and decide what to do next, but adds sequential token generation cost before every action.
For small, on-device models with limited reasoning capacity, lengthy texts and repeated deliberations may waste the generation cost without providing useful guidance for generating better actions.
An efficient agent needs reasoning that guides it to choose the desired action with minimal repeating cost of generating reasoning traces.

Existing approaches reduce this overhead by prompting concise traces or rewarding shorter reasoning~\citep{xu2025chain,arora2026training}.
These approaches reduce the amount of text generated, yet keep the sequential dependency between its tokens during autoregressive decoding.
We investigate an orthogonal direction: changing the representation of reasoning itself.
We observed that in multimodal search, entities, queries, and observations vary across questions, but the high-level reasoning for taking the next action often recurs.
For example, finding the location of an upcoming AI conference and the year LEGO was founded can share the same rationale: the entity is known but the requested information is missing, so search for the entity name and the attribute of interest.
Previous work on thought templates~\citep{yang2024buffer,ouyang2026reasoningbank} reported a similar finding, showing the value of reusing reasoning knowledge across problems.
Therefore, the required reasoning space for search agents does not span the entire natural language space.
Instead, the model can achieve a decent success rate with a few reasoning candidates to choose from.

Motivated by this, we introduce \textbf{Selection-based Structured Reasoning (SSR)}, a framework that formulates reasoning as \textit{selection} among reusable natural-language candidates instead of open-ended freeform generation.
At each turn, the agent selects a reasoning candidate conditioned on its interaction history, then uses the selected reasoning to generate the next action and its arguments (e.g., a text search with a question-specific query).
This design separates reusable reasoning patterns from instance-specific execution: the selected candidate provides the rationale for the next step, while action generation supplies the details needed to realize it.
The agent then adapts its actions to the current context without generating freeform reasoning traces at any stage.

\begin{figure*}[t]
  \centering
  \includegraphics[width=\textwidth]{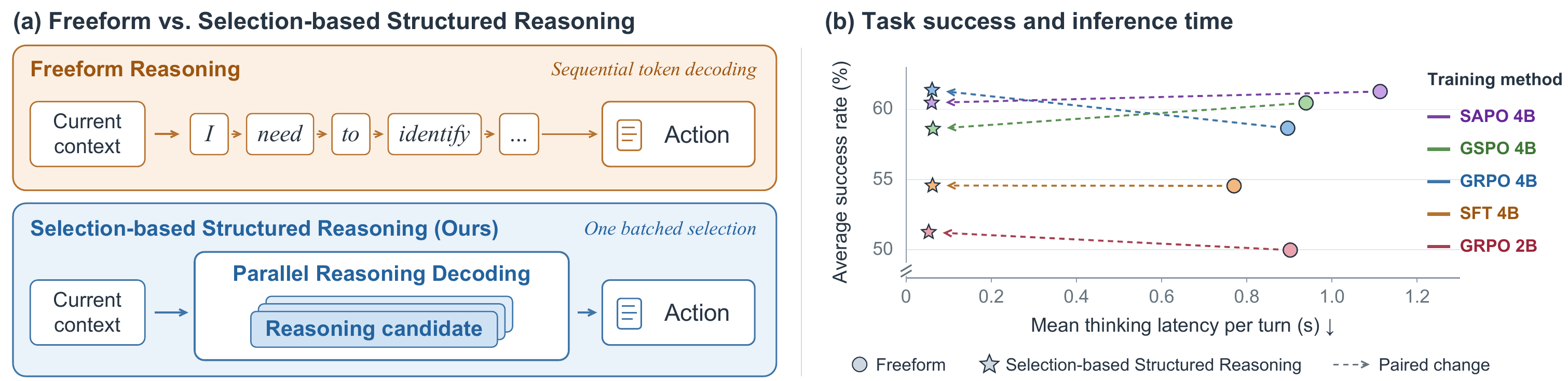}
  \caption{\textbf{SSR significantly reduces inference latency.} \textbf{Left:} Parallel reasoning decoding replaces autoregressive reasoning generation. \textbf{Right:} Across four training objectives and two model sizes, SSR achieves comparable or higher success rates while reducing mean reasoning latency per turn by more than $90\%$. Mean total model inference latency per question is reduced by $28$--$54\%$ (Table~\ref{tab:reasoning-comparison}).}
  \label{fig:teaser}
  \vspace{-5mm}
\end{figure*}

This representation enables \textbf{parallel reasoning decoding} (Figure~\ref{fig:teaser}).
We use the language model directly as a candidate classifier, without any additional head, to score each reasoning candidate by its length-normalized likelihood conditioned on the current context.
This leverages the language understanding capability of the model to match the context to the most suitable reasoning.
Since all candidate tokens are pre-specified, teacher forcing computes token likelihoods in parallel within each reasoning candidate, and candidates are also scored concurrently sharing the same KV cache.
The model then samples a reasoning candidate based on its score, and the harness inserts the selected reasoning into the model context as natural-language guidance for the next action.

We evaluate SSR across seven multimodal search benchmarks. Our 4B SSR with GRPO model achieves an average success rate of $61.37\%$, matching the performance of leading trained 4B search agents ($61.25\%$). 
Across GRPO~\citep{shao2024deepseekmath}, GSPO~\citep{zheng2025gspo}, SAPO~\citep{gao2025sapo}, and SFT, the SSR agents achieve comparable task performance while significantly reducing model inference latency compared to the freeform reasoning counterpart.
Against four representative efficient reasoning baselines, SSR achieves a higher success rate, lower model inference time, and more consistent latency.

In summary, our contributions are three-fold:
\vspace{-0.2em}
\begingroup
\setlength{\leftmargini}{1.5em}
\begin{itemize}\itemsep0pt
  \item \textbf{Reasoning by selection rather than generation.} We introduce Selection-based Structured Reasoning (SSR), which replaces freeform reasoning  with selection among reusable natural language candidates conditioned on the current context.
  \item \textbf{Parallel reasoning decoding.} We use the language model itself to score reasoning candidates by their length-normalized likelihoods. This approach requires neither an additional classification head nor an SFT warm start.
  Teacher forcing enables parallel scoring across all tokens within and between candidates, eliminating autoregressive generation of reasoning traces.
  \item \textbf{Competitive performance with substantial efficiency gains.} Across GRPO, GSPO, SAPO, and SFT, our 4B agents maintain competitive performance while reducing mean reasoning latency per turn by over $90\%$ and mean model latency per question by $28$–$54\%$.
\end{itemize}
\endgroup

%% file: section/2_related.tex
\section{Related Work}
\subsection{Multimodal Search Agents}
Prior work trained multimodal agents to interleave perception, retrieval, and tool use.
In multimodal search, MMSearch-R1~\citep{wu2025mmsearch} and DeepMMSearch-R1~\citep{narayan2025deepmmsearch} trained models to use text and image search, while Visual-ARFT~\citep{liu2025visual} and DeepEyes~\citep{zheng2026deepeyes} added image operations such as cropping and zooming.
DeepEyesV2~\citep{hong2025deepeyesv2} combined image operations with code execution and web search.
SenseNova-MARS~\citep{chng2025sensenova}, TAPO~\citep{dong2026tapo}, and WebWatcher~\citep{geng2025webwatcher} investigated multi-turn reinforcement learning, tool-aware credit assignment, and deep-research behavior.
More recent systems extended the scale and supervision of this recipe: VSearcher trained long-horizon search over text, images, and webpages~\citep{zhang2026vsearcher}, while ProMMSearchAgent introduced process-oriented rewards in a static sandbox before transfer to live search~\citep{yan2026prommsearch}.
These approaches focused on training-data mixtures, gradient routing, and credit assignment for performance improvements, while we explore an orthogonal direction that improves inference efficiency.

\subsection{Efficient Reasoning for MLLM Agents}
Efficient-reasoning methods controlled model deliberation through token budgets, early stopping, compressed chains of thought, or adaptive test-time computation~\citep{han2025tokenbudget,ma2025nothinking,sui2025stop}.
Agentic settings add another resource axis: the model must decide not only how long to think, but also whether to spend the next unit of budget on retrieval, interaction, or answer commitment.
Inference-time budget control for search agents explicitly modeled this trade-off~\citep{fang2026budgetsearch}.
Closest to our efficiency goal, HyperEyes~\citep{li2026hypereyes} treated inference efficiency as a
first-class training objective: it dispatched multiple grounded queries concurrently within a round and
rewarded trajectories that avoided superfluous tool calls, reducing the \emph{number} of interaction rounds,
whereas we reduce the \emph{cost of the reasoning span} within each round.
Grammar-constrained decoding enforced admissible output forms~\citep{geng2023grammar}, with engines such as XGrammar reducing the cost of constraint checking~\citep{dong2025xgrammar}.
SSR instead scores complete candidates in parallel and directly supplies the selected text as the reasoning trace.

\subsection{Reusable Reasoning and Agent Skills}

Prior work explored reusable knowledge through thought templates~\citep{yang2024buffer}, context-aware guidelines~\citep{fu2024autoguide}, reasoning memories~\citep{ouyang2026reasoningbank}, and task workflows~\citep{wang2025workflow}.
Given a library of reasoning candidates, SSR directly uses the acting MLLM's internal semantic knowledge to select among them through context-conditioned likelihoods, without any auxiliary knowledge-management or retrieval system.
Trajectory-level approaches such as StraTA~\citep{xue2026strata} generated a strategy from the initial task state and kept that strategy fixed throughout the interaction.
SSR instead selects a reasoning candidate at every turn based on the updated interaction history, allowing new observations to change the guidance for subsequent actions.
Programmatic approaches \citep{wang2023voyager, zheng2025skillweaver} encoded reusable skills as fixed, executable programs.
An SSR candidate instead supplies natural-language guidance that biases action generation without prescribing executable control flow.
This leaves concrete queries, tool arguments, and answers flexible, allowing the same reasoning candidate to support different actions across contexts without modifying the candidate itself.

%% file: section/3_method.tex
\section{Selection-Based Structured Reasoning}
\label{sec:method}

\subsection{Overview}
\label{sec:overview}

We introduce \textbf{Selection-based Structured Reasoning (SSR)}, which replaces freeform reasoning in ReAct agents~\citep{yao2022react} with \textit{selection} among reusable reasoning candidates.

At turn $t$, the interaction history of an agent is $h_t=[x,z_0,a_0,o_0,\ldots,z_{t-1},a_{t-1},o_{t-1}]$, where $x$ is the initial user request, $z_i$ is a reasoning trace, $a_i$ is an action, and $o_i$ is the observation returned by an external tool. A tool call extends the history with a new observation, while a final answer terminates the trajectory. Conventionally, an MLLM with distribution $p_\theta$ generates a freeform reasoning trace $z_t\sim p_\theta(\cdot\mid h_t)$ before generating the action $a_t\sim p_\theta(\cdot\mid h_t,z_t)$. Since $z_t$ is unconstrained natural-language text, its tokens must be decoded autoregressively, resulting in substantial inference latency.

\begin{figure*}[t]
  \centering
  \includegraphics[width=\textwidth]{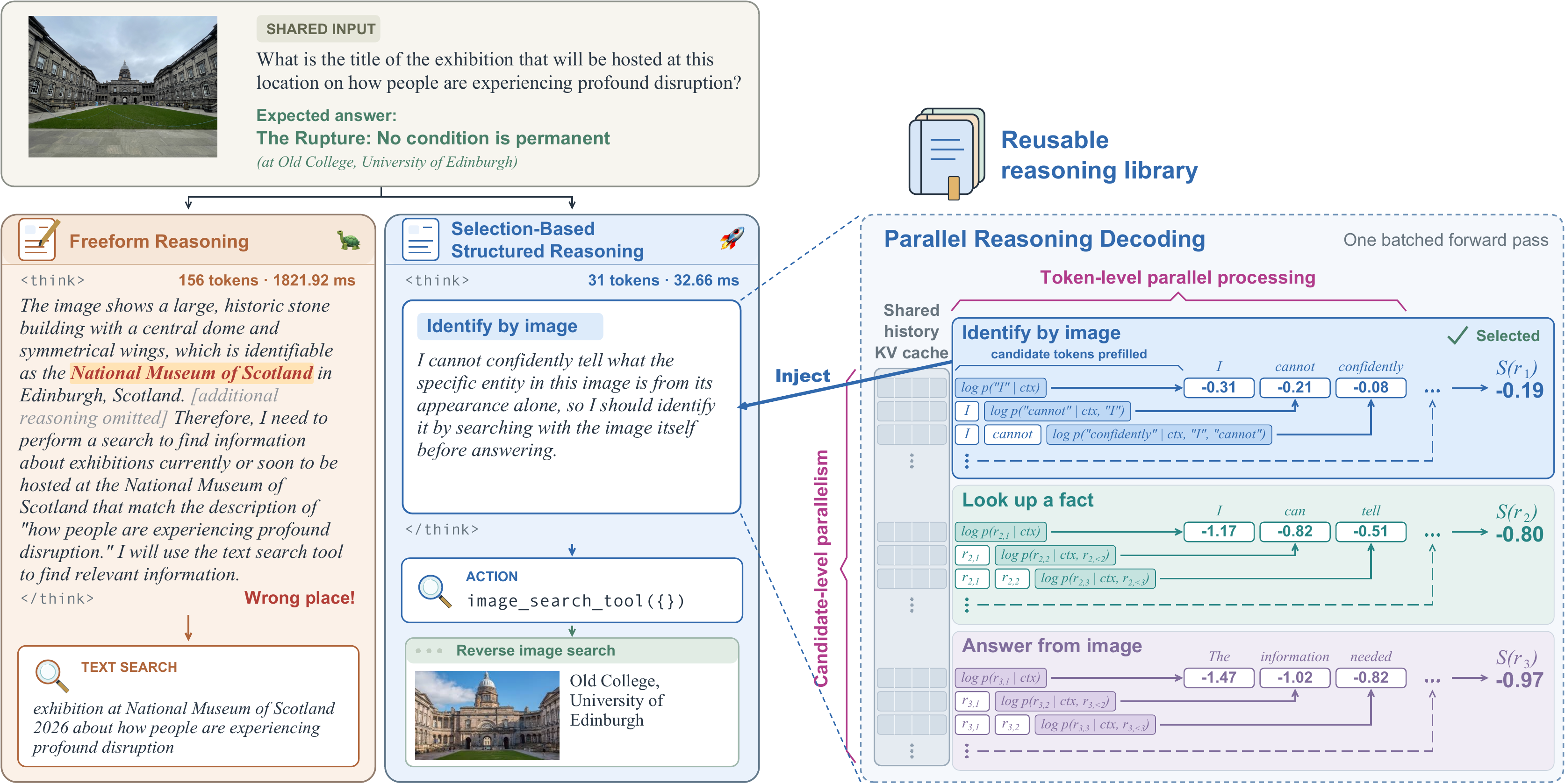}
  \caption{\textbf{Overview of Selection-based Structured Reasoning.}
  Freeform agents generate token-by-token reasoning before each action.
  SSR instead selects from a library of reusable natural-language reasoning candidates.
  Parallel reasoning decoding scores every token in each candidate using teacher-forced prefill, and the selected candidate guides context-specific action generation.
  }
  \label{fig:overview}
  \vspace{-3mm}
\end{figure*}

To eliminate this sequential decoding cost, SSR selects $z_t$ from reusable natural-language candidates conditioned on $h_t$, instead of generating it token by token (Figure~\ref{fig:overview}).
SSR contains a \textit{reasoning library} $\mathcal R=\{r_1,\ldots,r_N\}$, which consists of short natural-language candidates.
Each candidate is a reusable reasoning trace that describes what information is available or missing in the current context and what step to take next.
\textit{Parallel reasoning decoding} uses the language model itself to score the full text of each candidate conditioned on $h_t$ and select one as $z_t$.
Since all candidate texts are specified in advance, their token likelihoods can be computed with teacher-forced prefill and different candidates can be scored in parallel.
The selected reasoning is then inserted into the context to guide instance-specific action generation, such as a search query, crop location, or final answer.
This formulation leverages the semantics prior in the language decoder for reasoning selection, and makes use of the natural-language guidance for the action generation.

\subsection{Parallel Reasoning Decoding}
\label{sec:selector}

\paragraph{Scoring and selection.}
We define a scoring function to measure how suitable each reasoning candidate is for the current context.
Specifically, for a candidate $r_i$ that contains $K_i$ tokens $(r_{i,1},~r_{i,2},~\ldots,~r_{i,K_i})$,  we compute its length-normalized log-likelihood given context $h_t$ as:
\begin{equation}
S_\theta(r_i\mid h_t)
=\frac{1}{K_i^{\alpha}}\sum_{j=1}^{K_i}
\log p_\theta(r_{i,j}\mid h_t,r_{i,<j})
\label{eq:strategy-score}
\end{equation}
where $\alpha\in[0,1]$ controls length normalization.
We use $\alpha=1$ to compare candidates by their mean token log-likelihood, reducing the length bias from summing negative log-probabilities, which otherwise favors the shortest candidate. %
Applying a softmax gives the reasoning-selection policy $q_\theta(\cdot\mid h_t)$, a categorical distribution over $\mathcal R$:
\begin{equation}
q_\theta(r_i\mid h_t)
=\frac{\exp(S_\theta(r_i\mid h_t)/\tau)}
{\sum_{k=1}^{N}\exp(S_\theta(r_k\mid h_t)/\tau)}
\label{eq:strategy-policy}
\end{equation}
where $\tau>0$ is the selection temperature.
This represents a context-dependent preference over the reasoning candidates.
The model samples $~r_t\sim q_\theta(\cdot\mid h_t)$, inserts the selected text as the reasoning trace, and generates the action from $~p_\theta(\cdot\mid h_t,r_t)$.
Scoring and action generation use the same model parameters and system prompt, without the need for any extra components.
This mechanism also constrains the reasoning of the model to be in the library, regardless of its initial distribution.

\paragraph{Latency improvement through parallelism.}
Since all reasoning candidates are pre-specified, we exploit both candidate-level and token-level parallelism to reduce reasoning latency.
At the token level, since the prefix $r_{i,<j}$ (Eq.~\ref{eq:strategy-score}) is part of the given candidate, we can prefill all of the tokens in one reasoning candidate through teacher forcing and score the log-probabilities of all tokens in one batched forward pass. 
All tokens in the reasoning text will be processed concurrently parallel within each attention layer.
At the candidate level, each $S_\theta(r_i\mid h_t)$ can be computed concurrently given the same history KV cache for $h_t$.
Throughout this reasoning process, the model does not need to generate any new token.

Next, we characterize the latency advantage of parallel reasoning decoding in terms of computational depth.
Consider a model with fixed number of layers, token dimensions, and vocabulary size.
Let $H$ be the number of tokens in the history context; $N$ be the number of candidates in the library; $K_\text{max} = \max_i K_i$ be the maximum sequence length of the candidates; and $L$ be the number of tokens in a freeform reasoning trace generated by autoregressive decoding.
Then we can compare the length of the critical paths:
\begin{equation}
\begin{aligned}
\mathcal D_{\mathrm{freeform}}
&=O\!\Bigl(\sum_{\ell=0}^{L-1}\log(H+\ell)\Bigr)=O\!\left(L\log(H+L)\right) \\
\mathcal D_{\mathrm{SSR}}
&=O\!\Bigl(
\underbrace{\log(H+K_{\max})}_{\text{scoring the longest candidate}}
+
\underbrace{\log N}_{\text{selecting among the candidates}}
\Bigr)
\end{aligned}
\label{eq:reasoning-parallel-depth}
\end{equation}
where applying attention over $m$ tokens has $O(\log m)$ parallel depth due to the softmax and weighted sum operations.
The difference in time complexity highlights the latency advantage of the proposed selection method, which removes the factor of $L$ through token-level parallelism.

\subsection{SSR Training}
\label{sec:training}

SSR provides a general alternative to freeform reasoning under both supervised fine-tuning (SFT) and reinforcement learning (RL) objectives. At each turn, the policy makes a categorical reasoning selection and then autoregressively generates the action and its arguments. The reasoning-selection and action-generation objectives share the same model parameters and can be optimized jointly in an end-to-end manner.

\paragraph{Supervised fine-tuning.}
Given context $h_t$, ground truth reasoning candidate $z_t$ from the library, and ground truth action $a_t$ (tool call or answer), we define the SFT loss as a weighted sum of the categorical selection loss and the token-level action loss (modulated by weighting factor $\lambda^\text{SFT}_{\mathrm{reason}}$):
\begin{equation}
\mathcal L_{\mathrm{SFT}}
=-\sum_t\left[
\lambda^\text{SFT}_{\mathrm{reason}}\log q_\theta(z_t\mid h_t)
+\frac{1}{|a_t|}\sum_{k=1}^{|a_t|}
\log p_\theta(a_{t,k}\mid h_t,z_t,a_{t,<k})
\right]
\label{eq:sft}
\end{equation}

\paragraph{GRPO-style reinforcement learning.}
Unlike many prior methods ~\citep{hong2025deepeyesv2} that require an SFT warm-up to teach the base model a specific way of reasoning, SSR can be trained directly with RL. 
This is attributable to reasoning selection restricting the model to candidates in the library.
Compared with freeform GRPO~\citep{shao2024deepseekmath}, the main difference is to replace the token-level importance ratios for generated reasoning with a single categorical ratio for reasoning selection.
The action tokens retain their usual importance ratios:
\begin{equation}
\rho^{\mathrm{reason}}_t
=\frac{q_\theta(z_t\mid h_t)}{q_{\mathrm{old}}(z_t\mid h_t)}
\qquad
\rho^{\mathrm{action}}_{t,k}
=\frac{p_\theta(a_{t,k}\mid h_t,z_t,a_{t,<k})}
{p_{\mathrm{old}}(a_{t,k}\mid h_t,z_t,a_{t,<k})}
\label{eq:reason-action-ratios}
\end{equation}
where $q_{\mathrm{old}}$ and $p_{\mathrm{old}}$ use the rollout parameters $\theta_{\mathrm{old}}$.
The inserted reasoning text contributes to its \textit{single} categorical term. 
After a candidate is selected, the harness inserts its entire token sequence into the model context as a single block.

Specifically, for each question $x$, we sample multiple trajectories to form group $G$ by selecting $z_t\sim q_{\mathrm{old}}(\cdot\mid h_t)$ and generating the action autoregressively from $p_{\mathrm{old}}(\cdot\mid h_t,z_t)$ at each turn.
For trajectory $g$ in the group, its group-normalized advantage is $\widehat A_g=(R_g-\overline R)/\sigma_R$, where $R_g$ is its reward, $\overline R$ and $\sigma_R$ are the reward mean and standard deviation within the group.
After computing the ratios in Eq.~\ref{eq:reason-action-ratios}, they enter the clipped surrogate $\ell_\epsilon(\rho,A)=\min\{\rho A,\operatorname{clip}(\rho,1-\epsilon,1+\epsilon)A\}$.
And the final objective combines the reasoning-selection and action-token terms:
\begin{equation}
\mathcal J_{\mathrm{GRPO}}(\theta)
=\mathbb E_{g \in G}\!\left[
\displaystyle\sum_{t=0}^{T_g-1}
\left[\gamma^{\mathrm{RL}}_{\mathrm{reason}}\ell_\epsilon(\rho^{\mathrm{reason}}_{g,t},\widehat A_g)
+\sum_k\ell_\epsilon(\rho^{\mathrm{action}}_{g,t,k},\widehat A_g)\right]
\right]-\beta\mathcal K_{\mathrm{ref}}(\theta)
\label{eq:grpo-objective}
\end{equation}
where $T_g$ is the number of turns, and $\mathcal K_{\mathrm{ref}}$ is the reference-policy KL regularizer on generated action-token distributions, weighted by $\beta$.
The same reasoning-selection and action-generation components can also be trained with other GRPO variants, such as GSPO and SAPO.
Appendix~\ref{app:selector-gradient} gives the variant-specific objectives and practical selector-gradient approximation.

%% file: section/4_exp.tex
\section{Experiments}
\label{sec:experiments}

\subsection{Setup}
\label{sec:setup}

\paragraph{Datasets.}
Following prior work~\citep{chng2025sensenova,dong2026tapo}, we use VisualProbe~\citep{lai2026mini}, DeepEyes~\citep{zheng2026deepeyes}, and FVQA-train~\citep{wu2025mmsearch} for training,
and use MMSearch \citep{jiang2024mmsearch}, HR-MMSearch \citep{chng2025sensenova}, FVQA-test \citep{wu2025mmsearch}, InfoSeek \citep{chen2023infoseek}, SimpleVQA \citep{cheng2025simplevqa}, LiveVQA \citep{fu2025livevqa}, and MAT-Search \citep{liu2025visual} for evaluation.
Appendix~\ref{app:benchmarks} details the datasets used.

\paragraph{Implementation details.}
We use the 2B/4B Qwen3-VL~\citep{team2025qwen3vl} models for our experiments.
Following the aforementioned prior work, we use the same tool harness with reverse-image search, text search, and image cropping, along with the same LLM-as-judge for evaluation.
Our reasoning library contains six manually written reasoning candidates shared across benchmarks (details in Appendix~\ref{app:strategy-library}):
(1) doing an image search to identify the entity when unsure, (2) looking up specific information through a text search, (3) zooming into a specific part of the image for targeted inspection, (4) answering the question from the gathered evidence, (5) answering the question based on the image, and (6) answering with general knowledge.
Each of them covers a general case that the search agent may encounter.
Unless otherwise noted, we train each RL model in a single stage for one epoch with a learning rate of $10^{-5}$ using outcome and format rewards.
For simplicity, we set $\lambda^\text{SFT}_\text{reason}=\gamma^{\mathrm{RL}}_{\mathrm{reason}}=1$ and $\beta=10^{-3}$.
Appendix~\ref{app:training-details} contains additional details.

\paragraph{Evaluation metrics.}
We report success rate on all seven multimodal search benchmarks and inference efficiency.
For latency profiling, we sample $100$ questions per benchmark and measure per-turn reasoning latency (time to produce the reasoning trace) and per-question model latency (prefill, reasoning, action generation, and other model inference over the full trajectory).
We report mean and 95th-percentile latency. 
Environment-side processing, including tool execution and LLM-as-judge evaluation, is excluded as it introduces excessive model-independent variations.
We also report effective reasoning throughput, computed as the total number of reasoning tokens processed divided by the total reasoning time.
For SSR, this counts all candidate tokens being scored in parallel.
For autoregressive generation, it counts the tokens in the generated reasoning trace.
We run the profiling on an NVIDIA H100 GPU with an offline SGLang~\citep{zheng2024sglang} inference engine.

\subsection{SSR Achieves Competitive Task Performance}
\label{sec:downstream}

We first compare SSR with the state-of-the-art multimodal search agents.
Table~\ref{tab:mmsearch-main} shows that the 4B SSR model reaches $61.37\%$ average success rate, comparable to the best 4B SOTA model TAPO with GSPO at $61.25\%$ and outperforms 4B MMSearch-R1 and SenseNova-MARS baselines.
The 2B SSR agent reaches $51.26\%$, comparable to the zero-shot 8B model, showing the generalization of the approach to a smaller backbone.
These results show that selecting reusable reasoning can support competitive success rates in multimodal agentic search tasks. 

\begin{table}[!t]
\caption{
    \textbf{Success rate (\%) on multimodal search benchmarks.}
    SSR achieves comparable or better performance than state-of-the-art multimodal search agents of the same size.
    \textbf{Bold} denotes the best result among 4B models; \underline{underlines} mark the second best.
    ($^*$ reproduced results).
  }
\label{tab:mmsearch-main}
\centering
\begingroup
\setlength{\tabcolsep}{2.6pt}
\renewcommand{\arraystretch}{1.03}
\footnotesize
\resizebox{\textwidth}{!}{%
\begin{tabular}{llcccccccc}
\toprule
\textbf{Model} & \textbf{Size} & \shortstack{\textbf{MM}\\\textbf{Search}} & \shortstack{\textbf{HR-MM}\\\textbf{Search}} & \shortstack{\textbf{FVQA}\\\textbf{test}} & \shortstack{\textbf{Simple}\\\textbf{VQA}} & \shortstack{\textbf{Live}\\\textbf{VQA}} & \shortstack{\textbf{MAT}\\\textbf{Search}} & \textbf{InfoSeek} & \textbf{AVG} \\
\midrule
\multicolumn{10}{l}{\textit{Agentic models: zero-shot (proprietary)}} \\
GPT-4o-mini~\citep{openai2024gpt4omini} & -- & 38.60 & 26.23 & 50.00 & 50.84 & 31.54 & 80.00 & 42.35 & 45.65 \\
GPT-4o~\citep{openai2024gpt4o} & -- & 49.12 & 30.16 & 66.34 & 63.67 & 40.09 & 76.67 & 59.55 & 55.09 \\
GPT-5~\citep{openai2025gpt5} & -- & 52.63 & 38.36 & 62.61 & 70.58 & 56.02 & 84.67 & 55.95 & 60.12 \\
Gemini-3-Flash~\citep{google2025gemini3flash} & -- & 62.57 & 41.64 & 64.89 & 67.92 & 48.06 & 82.67 & 61.10 & 61.26 \\
GPT-5.2~\citep{openai2025gpt52} & -- & 66.08 & 48.20 & 68.78 & 78.18 & 65.99 & 80.67 & 65.55 & 67.64 \\
Gemini-3-Pro~\citep{google2026gemini3pro} & -- & 74.27 & 48.52 & 72.61 & 75.91 & 59.69 & 86.00 & 66.45 & 69.06 \\
\midrule
\multicolumn{10}{l}{\textit{Agentic models: zero-shot (open-source)}} \\
Qwen3-VL-32B-Instruct~\citep{team2025qwen3vl} & 32B & 49.12 & 34.43 & 54.28 & 64.17 & 42.87 & 82.00 & 49.85 & 53.82 \\
Qwen2.5-VL-32B-Instruct~\citep{bai2025qwen25vl} & 32B & 49.71 & 33.44 & 52.22 & 65.15 & 42.17 & 81.33 & 50.10 & 53.45 \\
Qwen3-VL-8B-Instruct~\citep{team2025qwen3vl}  & 8B & 47.37 & 27.87 & 53.61 & 62.29 & 39.37 & 82.67 & 46.15 & 51.33 \\
Qwen2.5-VL-7B-Instruct~\citep{bai2025qwen25vl}  & 7B & 32.16 & 19.34 & 36.00 & 42.35 & 22.52 & 67.33 & 28.80 & 35.50 \\
Qwen3-VL-4B-Instruct~\citep{team2025qwen3vl}  & 4B & 43.86 & 25.57 & 50.50 & 63.57 & 37.34 & 79.33 & 43.25 & 49.06 \\
Qwen3-VL-2B-Instruct~\citep{team2025qwen3vl}  & 2B & 32.35 & 20.00 & 42.90 & 49.16 & 31.69 & 70.47 & 33.20 & 39.97 \\
\midrule
\multicolumn{10}{l}{\textit{Agentic models: trained ({\tiny $\ge 7B$})}} \\
Skywork-R1V4~\citep{zhang2025skywork} & 30B & 66.10 & -- & 67.20 & -- & -- & -- & -- & -- \\
SenseNova-MARS~\citep{chng2025sensenova} & 8B & 66.67 & 40.33 & 67.11 & 70.19 & 56.22 & 83.33 & 61.70 & 63.65 \\
TAPO + SAPO~\citep{dong2026tapo} & 8B & 67.21 & 40.66 & 69.89 & 73.64 & 59.71 & 82.67 & 62.70 & 65.21 \\
Qwen3-VL-8B-Instruct + SAPO~\citep{gao2025sapo} & 8B & 64.33 & 38.69 & 65.61 & 70.38 & 56.86 & 82.00 & 59.55 & 62.48 \\
Visual-ARFT~\citep{liu2025visual} & 7B & 34.50 & 24.92 & 41.72 & 42.45 & 25.40 & 74.00 & 37.95 & 40.13 \\
DeepMMSearch-R1~\citep{narayan2025deepmmsearch} & 7B & -- & -- & -- & 55.87 & -- & -- & 47.51 & -- \\
MMSearch-R1~\citep{wu2025mmsearch} & 7B & 53.80 & 20.33 & 58.40 & 57.40 & 48.40 & 74.00 & 55.10 & 52.49 \\
DeepEyesV2~\citep{hong2025deepeyesv2} & 7B & 63.70 & -- & 60.60 & 59.40 & -- & -- & 51.10 & -- \\
\midrule
\multicolumn{10}{l}{\textit{Agentic models: trained ({\tiny $\le 4B$})}} \\
MMSearch-R1$^*$~\citep{wu2025mmsearch} & 4B & 53.80 & 25.50 & \underline{67.20} & 68.50 & 50.60 & 78.00 & \underline{60.00} & 57.66 \\
SenseNova-MARS$^*$~\citep{chng2025sensenova} & 4B & 44.05 & 24.67 & 57.19 & 65.55 & 43.06 & 78.67 & 48.40 & 51.65 \\
TAPO + GRPO~\citep{dong2026tapo} & 4B & 54.97 & \textbf{33.11} & 65.56 & 68.31 & 54.05 & \underline{82.00} & 58.85 & 59.55 \\
TAPO + GSPO~\citep{dong2026tapo}  & 4B & \underline{60.82} & \underline{31.48} & 66.61 & 70.98 & \underline{56.14} & \textbf{83.33} & 59.40 & \underline{61.25} \\
TAPO + SAPO~\citep{dong2026tapo}  & 4B & 56.73 & \underline{31.48} & 67.17 & \textbf{71.27} & 55.36 & 81.33 & 59.95 & 60.47 \\
\bf SSR (ours) & 4B & \textbf{61.40} & 30.16 & \textbf{70.22} & \underline{71.08} & \textbf{57.19} & 78.67 & \textbf{60.90} & \textbf{61.37} \\
\bf SSR (ours) & 2B & 51.18 & 17.06 & 61.16 & 61.83 & 51.54 & 62.59 & 53.44 & 51.26 \\
\bottomrule
\end{tabular}%
}
\endgroup
\par\smallskip
\begin{minipage}{\textwidth}
\scriptsize
\end{minipage}
\end{table}

\subsection{Comparison Between SSR and Freeform Reasoning}
\label{sec:training-compatibility}

Table~\ref{tab:reasoning-comparison} compares freeform reasoning and SSR across GRPO, GSPO, and SAPO, three representative RL algorithms used in~\citet{dong2026tapo}, as well as SFT.
For the RL algorithms, we directly train the model with outcome and format rewards without a separate SFT stage.
For SFT training data construction, we use Gemini 3.1 Pro~\citep{google2026gemini31pro} to generate the ground truth trajectory using the reasoning candidates from the library.

\begin{table}[!t]
\caption{\textbf{Success rate and inference efficiency across training objectives and model sizes.}
SSR achieves success rates comparable to those of freeform reasoning while reducing reasoning latency per turn by over 90\% and model latency per question by 28–54\%.
\textit{Italic} percentages show relative changes from the paired freeform model. 
}
\label{tab:reasoning-comparison}
\label{tab:algorithms}
\centering
\begingroup
\setlength{\tabcolsep}{3pt}
\renewcommand{\arraystretch}{1.12}
\scriptsize
\resizebox{\textwidth}{!}{%
\begin{tabular}{llcccc}
\toprule
\textbf{Training Objective} &
\textbf{Size} &
\shortstack{\textbf{Success}\\\textbf{Rate (\%)} $\uparrow$} &
\shortstack{\textbf{Reasoning Latency}\\\textbf{per Turn (s)} $\downarrow$} &
\shortstack{\textbf{Model Latency}\\\textbf{per Question (s)} $\downarrow$} &
\shortstack{\textbf{Effective Reasoning}\\\textbf{Throughput (toks/s)} $\uparrow$} \\
\midrule
GRPO-freeform & 4B & 58.65 & 0.896 & 5.456 & 72.7 \\
GRPO-SSR & 4B & 61.37 {\scriptsize\textcolor{green!45!black}{\textit{(+4.6\%)}}} & 0.061 {\scriptsize\textcolor{green!45!black}{\textit{(-93.2\%)}}} & 2.514 {\scriptsize\textcolor{green!45!black}{\textit{(-53.9\%)}}} & 3029.9 {\scriptsize\textcolor{green!45!black}{\textit{(+4070.2\%)}}} \\
\midrule
GSPO-freeform & 4B & 60.45 & 0.939 & 5.195 & 71.9 \\
GSPO-SSR & 4B & 58.60 {\scriptsize\textcolor{red!75!black}{\textit{(-3.1\%)}}} & 0.063 {\scriptsize\textcolor{green!45!black}{\textit{(-93.3\%)}}} & 3.706 {\scriptsize\textcolor{green!45!black}{\textit{(-28.7\%)}}} & 2526.2 {\scriptsize\textcolor{green!45!black}{\textit{(+3414.6\%)}}} \\
\midrule
SAPO-freeform & 4B & 61.26 & 1.113 & 6.145 & 69.4 \\
SAPO-SSR & 4B & 60.46 {\scriptsize\textcolor{red!75!black}{\textit{(-1.3\%)}}} & 0.060 {\scriptsize\textcolor{green!45!black}{\textit{(-94.6\%)}}} & 2.862 {\scriptsize\textcolor{green!45!black}{\textit{(-53.4\%)}}} & 3215.3 {\scriptsize\textcolor{green!45!black}{\textit{(+4535.0\%)}}} \\
\midrule
SFT-freeform & 4B & 54.55 & 0.770 & 5.308 & 70.9 \\
SFT-SSR & 4B & 54.57 {\scriptsize\textcolor{green!45!black}{\textit{(+0.0\%)}}} & 0.062 {\scriptsize\textcolor{green!45!black}{\textit{(-91.9\%)}}} & 3.318 {\scriptsize\textcolor{green!45!black}{\textit{(-37.5\%)}}} & 2567.0 {\scriptsize\textcolor{green!45!black}{\textit{(+3520.5\%)}}} \\
\midrule
GRPO-freeform & 2B & 49.98 & 0.902 & 4.153 & 87.2 \\
GRPO-SSR & 2B & 51.26 {\scriptsize\textcolor{green!45!black}{\textit{(+2.6\%)}}} & 0.053 {\scriptsize\textcolor{green!45!black}{\textit{(-94.1\%)}}} & 2.125 {\scriptsize\textcolor{green!45!black}{\textit{(-48.8\%)}}} & 3302.6 {\scriptsize\textcolor{green!45!black}{\textit{(+3687.9\%)}}} \\
\bottomrule
\end{tabular}%
}
\endgroup
\end{table}

\paragraph{Competitive success rates across objectives and model sizes.}
SSR improves average success over freeform GRPO at both model sizes and matches freeform SFT.
It shows modest decreases under GSPO and SAPO. We hypothesize this comes from the selector-gradient approximation (Appendix~\ref{app:gspo-sapo}), which becomes less accurate across multiple gradient updates per rollout.

\paragraph{Lower reasoning latency across training objectives.}
SSR brings more than $90\%$ reduction to the per-turn freeform reasoning latency and reduces mean per-question model latency by $28$-$54\%$.
Scoring complete candidates in parallel removes the sequential generation of the reasoning trace, yielding higher effective reasoning throughput that matches the token processing speed during prefill.
These results show consistent latency benefits across different training methods.

\subsection{Inference Latency and Reasoning Throughput}
\label{sec:inference-efficiency}

To compare SSR with existing efficient-reasoning methods, we compare four baselines spanning three approaches: prompt-based compression with Chain-of-Draft~\citep{xu2025chain} and Sketch-of-Thought~\citep{aytes2025sketch}, learning shorter reasoning through efficiency-reward RL~\citep{arora2026training}, and optional reasoning trace decoding with a lightweight classification head in Probe \& Prefill~\citep{sun2026llm}.
We implement the efficiency baselines on the GRPO-freeform pipeline for a fair comparison and report the results in Table~\ref{tab:latency}.

\begin{table}[t]
\caption{\textbf{Success rate and efficiency measurement on the profiling subset.} Our approach yields the highest success rate while being significantly faster than the baselines.}
\label{tab:latency}
\centering
\begingroup
\setlength{\tabcolsep}{3pt}
\footnotesize
\resizebox{\textwidth}{!}{%
\begin{tabular}{lcccccc}
\toprule
\multirow{2}{*}{\textbf{Method}} &
\multirow{2}{*}{\shortstack{\textbf{Subset SR}\\\textbf{(\%)} $\uparrow$}} &
\multicolumn{2}{c}{\textbf{Reasoning Latency/Turn (s)}} &
\multicolumn{2}{c}{\textbf{Model Latency/Question (s)}} &
\multirow{2}{*}{\shortstack{\textbf{Effective Reasoning}\\\textbf{Throughput (toks/s)} $\uparrow$}} \\
\cmidrule(lr){3-4}\cmidrule(lr){5-6}
& & \textbf{Mean} $\downarrow$ & \textbf{p95} $\downarrow$
  & \textbf{Mean} $\downarrow$ & \textbf{p95} $\downarrow$ & \\
\midrule
MMSearch-R1-4B~\citep{wu2025mmsearch} & 51.59 & 1.235 & 1.892 & 5.601 & 7.502 & \underline{73.8} \\
SenseNova-MARS-4B~\citep{chng2025sensenova} & 49.57 & 0.960 & 1.533 & 5.289 & 10.397 & 72.5 \\
Chain-of-Draft~\citep{xu2025chain} & 54.06 & \underline{0.416} & \underline{0.798} & 3.657 & \underline{6.616} & 69.9 \\
Sketch-of-Thought~\citep{aytes2025sketch} & 56.07 & 0.543 & 1.054 & \underline{3.487} & 7.048 & 71.7 \\
Efficiency-reward RL~\citep{arora2026training} & \underline{56.28} & 0.756 & 1.273 & 4.638 & 8.675 & 68.9 \\
Probe \& Prefill~\citep{sun2026llm} & 55.49 & 0.890 & 1.564 & 5.219 & 11.369 & 65.8 \\
\midrule
\textbf{SSR (Ours)} & \textbf{58.36} & \textbf{0.061} & \textbf{0.071} & \textbf{2.514} & \textbf{4.620} & \textbf{3029.9} \\
\bottomrule
\end{tabular}%
}
\endgroup
\end{table}

\paragraph{Higher success rate with lower inference latency.}
SSR achieves the highest success rate and delivers approximately $7\times$ or greater speedup in reasoning on all baselines.
The gains extend to question-level: mean model latency is around $2.5s$, approximately $30\%$ lower than Sketch-of-Thought, the fastest baseline on this metric.
These results show that parallel reasoning decoding is a more effective than reasoning-compression methods at reducing latency while maintaining high success rate.

The speed advantage is even larger at the $95$th percentile.
SSR has a p95 reasoning time of $0.071s$, $10$ ms above its mean, whereas the p95 latency for the baselines range from $0.798$ to $1.892s$.
The tight mean-to-p95 gap indicates consistent reasoning latency in SSR, matching the predictable computation of scoring all candidates in parallel as opposed to freeform generation with varying lengths.
At the question level, p95 latency is $4.6s$, about a $30\%$ reduction from the $6.6s$ of the fastest baseline.

\subsection{Comparison Between Different Reasoning Methods}
\label{sec:decoding-ablation}

We ablate the generation methods and compare different ways to select reasoning from the library.
Table~\ref{tab:decoding} compares full-reasoning-text selection, index-only selection, and autoregressive reasoning generation.
These variants distinguish the effects of the reasoning selection mechanism in SSR.

\begin{table}[t]
\caption{
\textbf{Comparison between different reasoning methods.}
Selecting only the index degrades performance, while autoregressive decoding without SFT fails to limit its reasoning to the reasoning candidates in the library.
}
\label{tab:decoding}
\centering
\begingroup
\setlength{\tabcolsep}{3pt}
\footnotesize
\resizebox{\textwidth}{!}{%
\begin{tabular}{lcccccccc}
\toprule
\textbf{Reasoning Method} & \shortstack{\textbf{MM}\\\textbf{Search}} & \shortstack{\textbf{HR-MM}\\\textbf{Search}} & \shortstack{\textbf{FVQA}\\\textbf{test}} & \shortstack{\textbf{Simple}\\\textbf{VQA}} & \shortstack{\textbf{Live}\\\textbf{VQA}} & \shortstack{\textbf{MAT}\\\textbf{Search}} & \textbf{InfoSeek} & \textbf{AVG} \\
\midrule
\multicolumn{9}{l}{\textit{Parallel reasoning decoding}} \\
\quad Entire reasoning & \textbf{61.40} & \underline{30.16} & \textbf{70.22} & \textbf{71.08} & \textbf{57.19} & \underline{78.67} & \underline{60.90} & \textbf{61.37} \\
\quad Index only & \underline{60.82} & 22.30 & 67.56 & \underline{69.99} & 55.56 & 77.33 & 59.30 & 58.98 \\
\multicolumn{9}{l}{\textit{Autoregressive decoding}} \\
\quad With SFT & \bf 61.40 & 29.84 & \underline{69.22} & 69.10 & \underline{56.97} & 72.00 & \textbf{61.55} & 60.01 \\
\quad Without SFT ($^*$degraded to freeform) & 60.23 & \bf 33.11 & 67.28 & 68.41 & 54.67 & \bf 79.33 & 58.60 & \underline{60.23} \\
\bottomrule
\end{tabular}%
}
\endgroup
\end{table}

\paragraph{Reasoning representation for candidate selection.}
The index-only variant applies Eq.~\ref{eq:strategy-score} to the index of each candidate (e.g., \texttt{[1]}) rather than its full text, then relies on the harness to inject the selected candidate before action generation.
Both variants supply the same structured reasoning to the action generation, but index-only scoring selects without access to the semantics of the reasoning tokens.
This change lowers average success by $3.9\%$, affecting all seven benchmarks.
This comparison highlights the value of reasoning-candidate semantics for better context-dependent selection.

\paragraph{Selection vs. autoregressive generation.}
Autoregressive decoding can also generate the specified reasoning candidate when the model is trained to use the library entries.
With an SFT warm start followed by RL with a format reward, the 4B model reproduces reasoning-library entries and achieves average success comparable to parallel reasoning decoding.
However, without SFT, specifying in the prompt and applying a format reward during RL cannot enforce using only the reasoning candidates from the library, and the model instead generates freeform reasoning.
Thus, the autoregressive variant must learn to conform to the reasoning library, whereas our reasoning selection enforces this interface directly.
Additionally, parallel full-text scoring further removes the sequential generation of the selected reasoning candidate, combining effective reasoning with lower decoding latency.

\begin{figure}[t]
  \centering
  \captionsetup{font=small}
  \begin{minipage}[t]{0.32\textwidth}
    \centering
    \vspace{0pt}
    \captionsetup{type=figure}
    \parbox[t][1.84in][t]{\linewidth}{%
      \includegraphics[width=\linewidth]{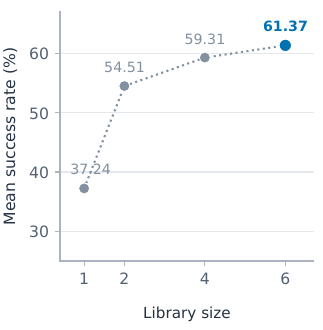}}
    \caption{\textbf{Reasoning library size.}
    Average success rate increases as the library size grows.}
    \label{fig:library-design}
  \end{minipage}\hfill
  \begin{minipage}[t]{0.655\textwidth}
    \centering
    \vspace{0pt}
    \captionsetup{type=figure}
    \parbox[t][1.84in][t]{\linewidth}{%
      \includegraphics[width=\linewidth]{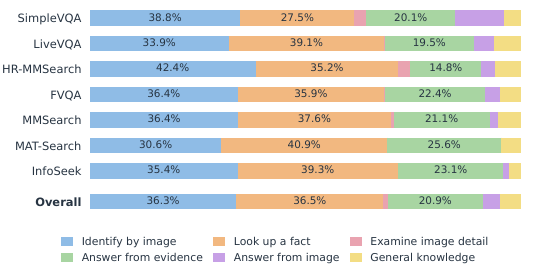}}
    \caption{\textbf{Reasoning selection distribution across benchmarks.}
    The same checkpoint exhibits different reasoning-selection distributions across benchmarks.
    }
    \label{fig:strategy-distribution}
  \end{minipage}
\end{figure}

\subsection{Reasoning Library Size and Selection Distribution}
\label{sec:library-ablation}

\paragraph{Effect of library size.}
Figure~\ref{fig:library-design} compares libraries with one, two, four, and six entries, ranging from a single generic candidate, ``I am a helpful assistant,'' to six more specific reasoning candidates.
Success rate across seven benchmarks increases monotonically from $37.24\%$ to $61.37\%$, showing that more detailed, granular reasoning candidates better guide learning for action generation.
The $17.27$ percentage point gain from one generic candidate to two candidates, answer or tool call, highlights the value of meaningful reasoning choices even in a small library.

\paragraph{Reasoning selection distribution.}
Figure~\ref{fig:strategy-distribution} shows that the same model yields different distributions of reasoning selection across different benchmarks.
Specifically, the simplest SimpleVQA has the highest selection rate for ``Answer from image'' and the lowest for ``Look up a fact'' and the high-resolution image centric HR-MMSearch has the highest selection rate for ``Examine image detail''.
These distribution differences suggest that SSR adapts its choices to question-specific information needs despite using a shared library.
These reasoning choices are reflected in action generation: in $99.9\%$ of turns, the generated action type matches the expected action type of the selected candidate.
The agent is indeed demonstrating behaviors that match the charasteristics of different benchmarks.

%% file: section/5_conclusion.tex
\section{Conclusion}

We presented \textbf{Selection-based Structured Reasoning (SSR)}, which reformulates reasoning in small multimodal search agents as selection rather than open-ended generation.
SSR captures recurring high-level reasoning as reusable natural-language candidates and selects among them based on their likelihoods under the current context.
Pre-specified candidates enable teacher-forced parallel scoring within and across candidates, eliminating token-by-token reasoning generation.
Across seven multimodal search benchmarks, SSR achieves competitive task performance while reducing per-turn reasoning latency by over $90\%$ and total per-question model inference latency by $28$-$54\%$.
More broadly, our results suggest that small search agents may not need to reason over the full language space at every step, and that organizing recurring reasoning into a compact, reusable space can enable more efficient multimodal agents.

%% file: section/6_statements.tex
\section*{Acknowledgements}
The authors sincerely thank Rosie Zhao, Yushu Pan, and Chaminda Bandara for their helpful discussions.

\section*{AI Use Statement}
We used generative AI tools to assist with drafting and revising the manuscript, checking mathematical claims and derivations, and locating and formatting references.
For the SFT experiments, Gemini 3.1 Pro generated reasoning traces and tool-call trajectories and mapped the retained traces to reasoning candidate labels, as described in Section~\ref{sec:experiments}.
The experimental pipeline also uses Qwen models to summarize retrieved content and judge open-ended answers, as detailed in Appendix~\ref{app:training-details}.
The authors take responsibility for the final text, mathematical claims, experimental results, and other AI-assisted content.

\section*{Reproducibility Statement}
Appendix~\ref{app:structured-reasoning-algorithm} lists the pseudocode for GRPO training and inference with SSR.
Appendix~\ref{app:benchmarks} documents the training data and the evaluation benchmarks.
Appendix~\ref{app:strategy-library} provides the reasoning library.
Appendix~\ref{app:training-details} describes the training data, hyperparameters, interaction limits, action space, scoring protocol, and rewards.

%% file: appendix/9_inference_algorithm.tex
\section{Pseudocode for SSR Inference and GRPO Training}
\label{app:structured-reasoning-algorithm}

\subsection{Inference}

Algorithm~\ref{alg:structured-reasoning-agent} implements SSR through SGLang's \texttt{generate} interface.\footnote{\url{https://docs.sglang.io/docs/basic_usage/sampling_params}}
The running context $h_t$ contains the complete interaction before turn $t+1$.
Each of the $N\geq1$ reasoning candidates is a nonempty token sequence including its reasoning delimiters; all inputs include the required chat delimiters and assistant prefixes.

\paragraph{Model interface and cache.}
Action sampling parameters $\eta$ specify the token limit (e.g., \texttt{max\_new\_tokens=8192}), temperature, and stopping conditions, with one completion per input.
While $h_t$ stores tokens, the engine-owned $\mathcal C_{\mathrm{KV}}$ stores computed model states; KV tensors are not endpoint arguments or return values.
New observations are cached when processed by the next model call.
For multimodal inputs, the adapter supplies matching \texttt{image\_data} per request row and tracks expanded token positions; these payloads are omitted from the pseudocode.

\paragraph{Candidate scoring.}
SGLang accepts variable-length candidate sequences without caller padding.
With \texttt{logprob\_start\_len}$=L_t-1$, the last history token forms a boundary, leaving the final $K_i$ input-token log-probabilities aligned with $r_i$.
Candidates are teacher-forced under causal attention and do not attend to one another.
Scores follow Eq.~(\ref{eq:strategy-score}), with $\alpha=1$ by default; selection follows Eq.~(\ref{eq:strategy-policy}) for $\tau>0$.
The selection temperature $\tau$ is independent of the action sampling settings $\eta$.

\begin{algorithm}[!t]
\caption{SSR inference using SGLang's \texttt{generate} interface}
\label{alg:structured-reasoning-agent}
\small
\begin{algorithmic}[1]
\REQUIRE Tokenized system prompt $s$ and initial question $q$; reasoning library $\mathcal R=\{r_1,\ldots,r_N\}$; persistent SGLang engine $\pi_\theta$ with prefix caching; harness $\mathcal H$; normalization exponent $\alpha\in[0,1]$; selection temperature $\tau\geq0$; action sampling parameters $\eta$; turn limit $T_{\max}\geq1$
\ENSURE $(h,\delta)$: final token context and stopping status
\item[] \textbf{Conventions:} $\Vert$ concatenates tokens; response indexing is zero-based; $[-K:]$ takes the last $K$ entries. Each \texttt{generate} call reuses and updates the engine-owned KV cache $\mathcal C_{\mathrm{KV}}$.
\item[] \textbf{Validity:} A final answer must be parseable and nonempty; a tool request must name an allowed tool and satisfy its argument schema, according to $\mathcal H$.
\STATE $h_0\gets s\Vert q$; $K_i\gets|r_i|$ for $i=1,\ldots,N$ \hfill $\triangleright$ context tokens; candidate lengths
\STATE $g_0\gets\{\texttt{"max\_new\_tokens"}:0\}$ \hfill $\triangleright$ prefill-only sampling parameters
\STATE $\begin{aligned}[t]
  \pi_\theta.\texttt{generate}(&\texttt{input\_ids}=s,\\
  &\texttt{sampling\_params}=g_0,\\
  &\texttt{return\_logprob}=\texttt{False}) \qquad \triangleright\text{ warm the system-prompt cache}
\end{aligned}$
\FOR{$t=0,\ldots,T_{\max}-1$}
  \STATE $L_t\gets|h_t|$; $X\gets[h_t\Vert r_i]_{i=1}^N$ \hfill $\triangleright$ $N$ variable-length token sequences
  \STATE $\begin{aligned}[t]
    Y\gets\pi_\theta.\texttt{generate}(&\texttt{input\_ids}=X,\\
    &\texttt{sampling\_params}=g_0,\\
    &\texttt{return\_logprob}=\texttt{True},\\
    &\texttt{logprob\_start\_len}=L_t-1) \qquad \triangleright\text{ list of }N\text{ responses}
  \end{aligned}$
  \FOR{$i=1,\ldots,N$}
    \STATE $D_i\gets Y[i-1].\texttt{input\_token\_logprob}[-K_i:]$
    \STATE Verify $[d[1]\text{ for }d\in D_i]=r_i$ \hfill $\triangleright$ record: (log-probability, token ID, text)
    \STATE $\ell_i\gets[d[0]\text{ for }d\in D_i]$; verify all entries are finite \hfill $\triangleright$ $|\ell_i|=K_i$
    \STATE $S_{t,i}\gets K_i^{-\alpha}\sum_{j=0}^{K_i-1}\ell_i[j]$ \hfill $\triangleright$ candidate scores $S_t\in\mathbb R^N$
  \ENDFOR
  \IF{$\tau=0$}\label{line:ssr-select}
    \STATE $i_t\gets\min\arg\max_i S_{t,i}$
  \ELSE
    \STATE $i_t\sim\operatorname{Categorical}(\operatorname{softmax}(S_t/\tau))$
  \ENDIF
  \STATE $z_t\gets r_{i_t}$; $u_t\gets h_t\Vert z_t$ \hfill $\triangleright$ append selected reasoning to current context
  \STATE $\begin{aligned}[t]
    A\gets\pi_\theta.\texttt{generate}(&\texttt{input\_ids}=u_t,\\
    &\texttt{sampling\_params}=\eta,\\
    &\texttt{return\_logprob}=\texttt{False})
  \end{aligned}$
  \STATE $a_t\gets A.\texttt{output\_ids}$; $h_t^+\gets u_t\Vert a_t$ \hfill $\triangleright$ action tokens; extended context
  \IF{$a_t$ is a valid final answer}
    \RETURN $(h_t^+,\texttt{answer})$
  \ELSIF{$a_t$ is not a valid tool request}
    \RETURN $(h_t^+,\texttt{invalid-output})$
  \ENDIF
  \STATE $(o_t,\delta_t)\gets\mathcal H.\texttt{execute}(a_t)$ \hfill $\triangleright$ observation tokens; \texttt{continue} or error
  \STATE $h_{t+1}\gets h_t^+\Vert o_t$ \hfill $\triangleright$ observation includes the next assistant prefix
  \IF{$\delta_t\neq\texttt{continue}$}
    \RETURN $(h_{t+1},\delta_t)$
  \ENDIF
\ENDFOR
\RETURN $(h_{T_{\max}},\texttt{turn-limit})$
\end{algorithmic}
\end{algorithm}

\paragraph{Constructing the reasoning library.}
A straightforward procedure to construct the reasoning library is to sample training questions and successful freeform reasoning traces, group recurring information needs, and summarize them into a small set of reusable reasoning candidates.
Each candidate describes how to make progress without retaining question-specific entities or answers.
Ideally, the set of candidates should cover all states that the agent will encounter.
Future work could explore skill-evolution methods such as SkillOpt~\citep{yang2026skillopt} to construct and refine the reasoning library automatically from trajectory feedback.

\subsection{Grouped GRPO with Detached Competitors}
\label{app:ssr-grpo-pseudocode}

Algorithm~\ref{alg:ssr-grpo-training} optimizes Eq.~(\ref{eq:grpo-objective}) using the detached-competitor approximation in Eq.~(\ref{eq:approx-selector}).
It alternates one fresh rollout minibatch with one optimizer step.

\begin{algorithm}[!t]
\caption{Grouped GRPO for SSR with detached competitor scores}
\label{alg:ssr-grpo-training}
\small
\begin{algorithmic}[1]
\REQUIRE Parameters $\theta$; frozen reference $p_{\mathrm{ref}}$; system prompt $s$; question distribution $\mathcal D$; library $\mathcal R$; harness $\mathcal H$; reward function $f_{\mathrm{reward}}$; questions per minibatch $m\geq1$; group size $n\geq2$; rollout settings $\alpha$, $\tau>0$, $\eta$, $T_{\max}$; training hyperparameters $\epsilon$, $\gamma^{\mathrm{RL}}_{\mathrm{reason}}$, $\beta$, $\delta$; optimizer $\mathcal O$
\ENSURE Trained parameters $\theta$
\item[] \textbf{Conventions:} $\Vert$ concatenates tokens; arrays are zero-based and candidate indices one-based. Rollout records are detached. $N=|\mathcal R|$; $\operatorname{LSE}(u)=\log\sum_i e^{u[i]}$.
\item[] \textbf{Surrogate:} $\ell_\epsilon(\rho,A)=\min\{\rho A,\operatorname{clip}(\rho,1-\epsilon,1+\epsilon)A\}$.
\WHILE{the training budget remains}
  \STATE $\theta_{\mathrm{old}}\gets\operatorname{stopgrad}(\operatorname{copy}(\theta))$
  \STATE Synchronize $\pi_{\mathrm{old}}$ to $\theta_{\mathrm{old}}$; invalidate its weight-dependent caches
  \STATE Sample $m$ questions $X\sim\mathcal D^m$; $\mathcal M\gets[\,]$ \hfill $\triangleright$ current rollout minibatch
  \FOR{each question $x\in X$}
    \FOR{$g=1,\ldots,n$}
      \STATE Collect $d_g$ with Algorithm~\ref{alg:structured-reasoning-agent} using $\pi_{\mathrm{old}}$ and question $x$, with turn logging
      \STATE $R_g\gets f_{\mathrm{reward}}(x,d_g)$ \hfill $\triangleright$ reward, including non-answer stopping cases
    \ENDFOR
    \STATE $\mu\gets n^{-1}\sum_g R_g$; $\sigma\gets\sqrt{n^{-1}\sum_g(R_g-\mu)^2}$
    \FOR{$g=1,\ldots,n$}
      \STATE $d_g.\texttt{adv}\gets\operatorname{stopgrad}((R_g-\mu)/\max(\sigma,\delta))$; append $d_g$ to $\mathcal M$
    \ENDFOR
  \ENDFOR
  \STATE $J\gets0$ \hfill $\triangleright$ accumulated surrogate; $|\mathcal M|=mn$ trajectories
  \FOR{each trajectory $d\in\mathcal M$}
    \FOR{each turn $v\in d.\texttt{turns}$}
      \STATE $(h,z,a,i)\gets(v.\texttt{h},v.\texttt{z},v.\texttt{a},v.\texttt{index})$; $K\gets|z|$; $w\gets z\Vert a$
      \STATE $b\gets[\log p_\theta(w[k]\mid h\Vert w[:k])]_{k=0}^{|w|-1}$ \hfill $\triangleright$ teacher forcing; gradients enabled
      \STATE $s_{\mathrm{sel}}\gets K^{-\alpha}\sum_{j=0}^{K-1}b[j]$ \hfill $\triangleright$ current selected-candidate score
      \STATE $u\gets\operatorname{stopgrad}(v.\texttt{scores}/\tau)$; $u[i-1]\gets s_{\mathrm{sel}}/\tau$ \hfill $\triangleright$ $N$ selector logits
      \STATE $\widetilde b^{\mathrm{reason}}\gets u[i-1]-\operatorname{LSE}(u)$
      \STATE $\rho^{\mathrm{reason}}\gets\exp(\widetilde b^{\mathrm{reason}}-v.\texttt{logq})$ \hfill $\triangleright$ one scalar ratio
      \STATE $\rho^{\mathrm{action}}\gets\exp(b[K:]-v.\texttt{logp})$ \hfill $\triangleright$ $|a|$ token ratios, elementwise
      \STATE $J\gets J+\gamma^{\mathrm{RL}}_{\mathrm{reason}}\ell_\epsilon(\rho^{\mathrm{reason}},d.\texttt{adv})+\sum_k\ell_\epsilon(\rho^{\mathrm{action}}[k],d.\texttt{adv})$
    \ENDFOR
  \ENDFOR
  \STATE $\widehat{\mathcal K}\gets$ reference-policy KL estimate on generated action positions in $\mathcal M$
  \STATE $\mathcal L\gets-J/|\mathcal M|+\beta\widehat{\mathcal K}$ \hfill $\triangleright$ scalar loss; trajectory average
  \STATE $\mathcal O.\texttt{zero\_grad}()$; $\mathcal L.\texttt{backward}()$; $\mathcal O.\texttt{step}()$ \hfill $\triangleright$ one update from $\mathcal M$
\ENDWHILE
\RETURN $\theta$
\end{algorithmic}
\end{algorithm}

\paragraph{Rollout and training settings.}
Rollout inputs follow Algorithm~\ref{alg:structured-reasoning-agent}.
Training uses clipping threshold $\epsilon>0$, reasoning-selection weight $\gamma^{\mathrm{RL}}_{\mathrm{reason}}\geq0$, reference-policy KL weight $\beta\geq0$, and reward-standard-deviation floor $\delta>0$.
For trajectory logging, set \texttt{return\_logprob=True} on the action-generation request in Algorithm~\ref{alg:structured-reasoning-agent}.
Logged behavior probabilities and current action likelihoods must use the same sampling distribution.
Each trajectory record retains its turns, final context, observations, media references, and stopping status so $f_{\mathrm{reward}}$ can also score unsuccessful or truncated trajectories.

\paragraph{Turn records.}
Each turn record stores context \texttt{h}, selected reasoning \texttt{z}, action \texttt{a}, candidate index $i_t$ (\texttt{index}), and rollout score vector $S_t^{\mathrm{old}}$ (\texttt{scores}).
It also stores $v.\texttt{logq}=S^{\mathrm{old}}_{t,i_t}/\tau-\operatorname{LSE}(S_t^{\mathrm{old}}/\tau)$ and $v.\texttt{logp}\in\mathbb R^{|a_t|}$, the action-token log-probabilities from \texttt{output\_token\_logprobs} in token order.
Verify their token IDs against $a_t$ and retain exactly $|a_t|$ finite values.

\paragraph{Differentiable update.}
Recompute likelihoods with the differentiable model $p_\theta$; SGLang rollout outputs are detached.
Only the selected candidate's score receives gradients, through both the softmax numerator and denominator.
Teacher-forced turns may be batched with causal attention and true-length masks.
Action-token losses and reference-policy KL apply only to generated action positions; prompts, inserted reasoning, and observations remain visible as context.
Exclude padding from both attention and loss.

%% file: appendix/1_selector_gradient.tex
\section{Selector-Gradient Approximation}
\label{app:selector-gradient}

\subsection{Exact and Approximate Gradients}

\paragraph{Exact gradient.}
For turn $t$ with history $h_t$ and a selected candidate $z_t$ from the library $\mathcal R=\{r_1,\ldots,r_N\}$ using selection temperature $\tau > 0$, the normalized policy in Eq.~(\ref{eq:strategy-policy}) satisfies
\begin{equation}
\log q_\theta(z_t\mid h_t)=\frac{S_\theta(z_t\mid h_t)}{\tau}
-\log\sum_{i=1}^{N}\exp\left(\frac{S_\theta(r_i\mid h_t)}{\tau}\right)
\end{equation}
Differentiating gives
\begin{equation}
\nabla_\theta\log q_\theta(z_t\mid h_t)
=\frac{1}{\tau}\left[\nabla_\theta S_\theta(z_t\mid h_t)
-\sum_{i=1}^{N}q_\theta(r_i\mid h_t)\nabla_\theta S_\theta(r_i\mid h_t)\right]
\label{eq:exact-selector-gradient}
\end{equation}
With length-normalization exponent $\alpha=1$,
\begin{equation}
\nabla_\theta S_\theta(r_i\mid h_t)
=\frac{1}{K_i}\sum_{k=1}^{K_i}\nabla_\theta\log p_\theta(r_{i,k}\mid h_t,r_{i,<k})
\end{equation}
where $r_{i,<k}$ is the candidate prefix before token $k$.
The exact gradient requires both forward and backward computation for \textbf{all} candidates, which can be very expensive.

\paragraph{Detached competitors.}
Let $\theta_{\mathrm{old}}$ be the rollout parameters.
We recompute the selected score at the current parameters and hold the competing scores at their rollout values, giving the approximation (note that the highlighted $S_{\theta_{\mathrm{old}}}$ represents the key difference between exact and approximated computation).
\begin{equation}
\widetilde q_\theta(z_t\mid h_t)
=\frac{\exp(S_\theta(z_t\mid h_t)/\tau)}
{\exp(S_\theta(z_t\mid h_t)/\tau)+\sum_{r_i\ne z_t}\exp({\color{red} S_{\theta_{\mathrm{old}}}}(r_i\mid h_t)/\tau)}
\label{eq:approx-selector}
\end{equation}
This avoids backward computation through competing candidates:
\begin{equation}
\nabla_\theta\log\widetilde q_\theta(z_t\mid h_t)
=\frac{1-\widetilde q_\theta(z_t\mid h_t)}{\tau}\nabla_\theta S_\theta(z_t\mid h_t)
\label{eq:detached-gradient}
\end{equation}
At $\theta=\theta_{\mathrm{old}}$ during the first gradient update after the rollouts are collected, the selected probability is exact. 
After an update, the stored competitor scores can differ from their current values.

\subsection{SSR Rollouts Are On-Policy in GRPO}

During rollout collection, parallel reasoning decoding scores the reasoning candidates under the rollout parameters $\theta_{\mathrm{old}}$ and samples $z_t\sim q_{\mathrm{old}}(\cdot\mid h_t)$.
The harness inserts the selected candidate into the context, after which the model generates action tokens autoregressively from $p_{\mathrm{old}}(\cdot\mid h_t,z_t)$.
The retained behavior probabilities therefore correspond to the process that generated the trajectory: one categorical probability for the reasoning selection and token-level probabilities for the subsequent action.
Selecting from the reasoning library defines the support of the reasoning policy, while inserting the selected text is deterministic conditional on that selection.
Thus, rollout collection is on-policy with respect to the composite SSR behavior policy.

When a rollout batch is reused after parameter updates, the current policy differs from the behavior policy, as in standard GRPO.
The reasoning- and action-level importance ratios in Eq.~(\ref{eq:reason-action-ratios}) compare the current probabilities with the fixed probabilities recorded during rollout collection.
The detached-competitor approximation in Eq.~(\ref{eq:approx-selector}) concerns the current selector probability and its gradient during optimization; it does not change the policy used to collect the rollouts.
Each inserted reasoning span is consequently represented by one categorical selection term, while token-level action terms apply only to the autoregressively generated action tokens.

\subsection{GSPO and SAPO with Reused Rollouts}
\label{app:gspo-sapo}

\paragraph{Update rules.}
For trajectory $g$ in rollout group $G$, let $\mathcal U_g$ index its categorical reasoning selections and generated action tokens.
With $T_g$ turns and action-token sequence $a_{g,t}$ at turn $t$, define
$Z_g=|\mathcal U_g|=\sum_{t=0}^{T_g-1}(1+|a_{g,t}|)$.
For $u\in\mathcal U_g$, $\rho_{g,u}$ is the corresponding ratio in Eq.~(\ref{eq:reason-action-ratios}), and $\widehat A_g$ is the group-normalized trajectory advantage defined in Section~\ref{sec:training}.
GSPO~\citep{zheng2025gspo} forms the length-normalized sequence ratio
\begin{equation}
\overline\rho_g=\exp\left(\frac{1}{Z_g}\sum_{u\in\mathcal U_g}\log\rho_{g,u}\right)
\label{eq:gspo-sequence-ratio}
\end{equation}
and applies the clipped surrogate $\ell_\epsilon(\overline\rho_g,\widehat A_g)$ from Section~\ref{sec:training} at the sequence level, with clipping threshold $\epsilon$.
SAPO~\citep{gao2025sapo} instead applies a smooth surrogate $f_\kappa(\rho)\widehat A_g$ to each ratio $\rho>0$, with slope $\kappa>0$ chosen according to the sign of $\widehat A_g$:
\begin{equation}
f_\kappa(\rho)=\frac{4}{\kappa[1+\exp(-\kappa(\rho-1))]}
\qquad
f_\kappa'(\rho)=\frac{4\exp(-\kappa(\rho-1))}{[1+\exp(-\kappa(\rho-1))]^2}
\label{eq:sapo-surrogate}
\end{equation}
The derivative attenuates updates as the ratio moves away from one.
Both comparisons normalize by trajectory length and average across trajectories, whereas Eq.~(\ref{eq:grpo-objective}) sums the within-trajectory terms.
Inserted reasoning tokens, prompts, observations, and padding do not add to $Z_g$.

\paragraph{Second and subsequent updates.}
As discussed for Eq~\ref{eq:approx-selector}, from the second update onward, the rollouts are stale relative to the current policy.
During repeated updates on a collected batch, competitor scores remain fixed at their rollout values.
Differences from current competitor scores affect the reasoning ratios: these enter the GSPO sequence ratio in Eq.~(\ref{eq:gspo-sequence-ratio}) and the SAPO attenuation in Eq.~(\ref{eq:sapo-surrogate}). The selector gradient uses the approximation in Eq.~ (\ref{eq:detached-gradient}).

%% file: appendix/6_inference_cost.tex
\section{Detailed Inference Cost}
\label{app:inference-cost}

Let $H$ be the cached history length, $L$ the freeform reasoning length, $N$ the library size, and $K_i$ the length of candidate $r_i$.
All lengths count tokens, including scored formatting tokens, and $K_{\mathrm{tot}}=\sum_{i=1}^{N}K_i$ is the total number of candidate tokens scored.
Equation~\ref{eq:reasoning-parallel-depth} compares the parallel computational depth of freeform reasoning and SSR.
We distinguish this quantity from the number of FLOPs, KV-cache memory usage, and inference latency.

\paragraph{FLOPs and KV-cache memory usage.}
For a fixed Transformer with causal self-attention, let $\gamma_{\mathrm{tok}}$ denote the approximate FLOPs per token for linear projections, feed-forward networks, and the output projection, summed across layers.
Let $\gamma_{\mathrm{att}}$ denote the approximate FLOPs per query--key pair for the attention computation, also summed across layers and heads.
We count a multiplication and an addition as two FLOPs and retain the leading terms in the operation count.
Let $W_{\mathrm{free}}$ and $W_{\mathrm{SSR}}$ denote the approximate FLOP counts for freeform generation and SSR candidate evaluation, respectively.
With the shared history already stored in the KV cache, we have
\begin{align}
W_{\mathrm{free}} &\simeq \gamma_{\mathrm{tok}}L+\gamma_{\mathrm{att}}\left(HL+\frac{L(L-1)}{2}\right) \\
W_{\mathrm{SSR}} &\simeq \gamma_{\mathrm{tok}}K_{\mathrm{tot}}+\gamma_{\mathrm{att}}\left(HK_{\mathrm{tot}}+\frac{1}{2}\sum_{i=1}^{N}K_i(K_i-1)\right)
\label{eq:structured-work}
\end{align}
This represents self-attention where each candidate attends to the shared history and its own preceding tokens.
Reusing the KV cache of the common prefix avoids recomputing history representations~\citep{zheng2024sglang}.
For fixed reasoning and candidate lengths, the ratio of attention FLOPs approaches $K_{\mathrm{tot}}/L$ as $H$ increases.

Let $m_{\mathrm{KV}}$ denote KV-cache storage in bytes per token across all layers.
If the common prefix is stored once and all candidate suffixes are retained, the KV-cache memory usage is approximately $m_{\mathrm{KV}}(H+K_{\mathrm{tot}})$ bytes.
Storing a separate copy of the prefix for each candidate instead requires approximately $m_{\mathrm{KV}}(NH+K_{\mathrm{tot}})$ bytes.
With the history stored once, candidate-suffix KV storage grows linearly with $K_\text{tot}$; for fixed candidate length, it grows linearly with the number of candidates.

\paragraph{Inference latency.}
For an execution stage $s$, let $\Delta_s$ denote its duration in seconds, $F_s$ its FLOP count, and $B_s$ the total bytes read from and written to device memory.
Let $\Pi$ be the peak floating-point performance in FLOPs/s at the chosen precision, and $\mathcal B$ the peak device-memory bandwidth in bytes/s.
These resource limits imply
\begin{equation}
\Delta_s\geq\max\left\{\frac{F_s}{\Pi},\;\frac{B_s}{\mathcal B}\right\}
\end{equation}
Here, arithmetic intensity is $F_s/B_s$, measured in FLOPs per byte.

Let $\Delta_{\mathrm{free}}$ and $\Delta_{\mathrm{SSR}}$ denote inference latency for generating and selecting the reasoning text, respectively.
Define $v_{\mathrm{dec}}$ as the number of generated tokens per second and $v_{\mathrm{score}}$ as the number of candidate tokens scored per second during model execution.
Let $\delta_{\mathrm{free}}$ and $\delta_{\mathrm{SSR}}$ denote additional runtime overheads excluded from those rates, such as candidate selection on the CPU.
Then
\begin{equation}
\Delta_{\mathrm{free}}=\delta_{\mathrm{free}}+L/v_{\mathrm{dec}}
\qquad
\Delta_{\mathrm{SSR}}=\delta_{\mathrm{SSR}}+K_{\mathrm{tot}}/v_{\mathrm{score}}
\end{equation}
Then, for $L>0$ and positive rates, SSR reduces inference latency if and only if
\begin{equation}
\frac{K_{\mathrm{tot}}}{L}<\frac{v_{\mathrm{score}}}{v_{\mathrm{dec}}}
+\frac{v_{\mathrm{score}}}{L}\left(\delta_{\mathrm{free}}-\delta_{\mathrm{SSR}}\right)
\label{eq:structured-break-even}
\end{equation}
Assuming the overhead difference is negligible, on the same hardware platform, this condition reduces to 
\begin{equation}
\frac{K_{\mathrm{tot}}}{L} < \frac{v_{\mathrm{score}}}{v_{\mathrm{dec}}}
\end{equation}
To put this into perspective, for an NVIDIA RTX PRO 6000 GPU, generating 100 freeform reasoning tokens autoregressively takes around the same time of processing a 15,200-token library.

%% file: appendix/2_search_benchmarks.tex
\section{Training and Evaluation Datasets}
\label{app:benchmarks}

\subsection{Training Datasets}

We use the 3,859-example training mixture from TAPO/SenseNova-MARS~\citep{dong2026tapo,chng2025sensenova}, with the subset sizes and accuracy rewards shown in Table~\ref{tab:mmsearch-training-data}.
The reward implementation is described in Appendix~\ref{app:training-details}.

\paragraph{FVQA-train.}
FactualVQA~\citep{wu2025mmsearch} combines factual image--question--answer pairs requiring visual entity recognition or associated factual knowledge, with a mixture of search-required and search-free questions.
It trains the agent to identify when image or text retrieval is needed.

\paragraph{DeepEyes-4K.}
DeepEyes~\citep{zheng2026deepeyes} provides visual questions spanning fine-grained perception, chart understanding, and challenging reasoning, supporting learning to inspect relevant image regions.
Our training mixture includes both multiple-choice (MCQ) and non-MCQ subsets.

\paragraph{VisualProbe-train.}
VisualProbe~\citep{lai2026mini} contains visual search questions about small targets in high-resolution images with distractor objects.
These questions encourage repeated inspection and exploration of image regions.

\begin{table}[htbp]
\caption{Multimodal-search RL training data.}
\label{tab:mmsearch-training-data}
\centering
\footnotesize
\begin{tabular}{lrl}
\toprule
\bf Subset & \bf Samples & \bf Accuracy reward \\
\midrule
FVQA-train & 1,894 & LLM judge \\
DeepEyes-4K (non-MCQ) & 425 & LLM judge \\
Visual-Probe-train & 912 & LLM judge \\
DeepEyes-4K (MCQ) & 628 & Exact match \\
\midrule
Total & 3,859 & --\\
\bottomrule
\end{tabular}
\end{table}

\subsection{Evaluation Benchmarks}

We evaluate on seven publicly available multimodal search benchmarks that collectively span a broad range of retrieval difficulty, temporal freshness, and image resolution requirements.
We use the benchmark versions distributed in the SenseNova evaluation collection~\citep{chng2025sensenova}.

\paragraph{MMSearch} \citep{jiang2024mmsearch} contains 300 manually collected questions spanning 14
subfields, of which 171 are accompanied by an image; its News area covers events from May
to August 2024, testing the model's ability to ground visual observations in up-to-date factual knowledge.
We follow the standard split of 171 questions used for system-level evaluation.

\paragraph{HR-MMSearch} \citep{chng2025sensenova} is a knowledge-intensive benchmark comprising 305 questions paired with high-resolution (4K) images sourced from international news agencies (2025 events), spanning eight topical domains.
Approximately 60\% of questions are classified as hard, requiring three or more tool invocations to answer correctly.

\paragraph{FVQA-test} \citep{wu2025mmsearch} is a curated 1,800-sample test set consisting of three equal-sized subsets: FVQA-auto-vc (600 automatically constructed samples, each manually checked for correctness and disjoint from the training set), InfoSeek Human Split (600 samples with manually corrected annotations), and an expert-annotated subset (600 samples).

\paragraph{InfoSeek} \citep{chen2023infoseek} is a visual question answering dataset for
information-seeking questions that cannot be answered through commonsense reasoning alone. It is built
on visual entity recognition and grounded in Wikidata, targeting fine-grained factual
attributes that models do not reliably surface without retrieval.
We sample 2,000 instances from its test split.

\paragraph{SimpleVQA} \citep{cheng2025simplevqa} focuses on factual accuracy, combining post-2023 VQA instances with expert-curated image-question pairs across 1,013 English samples.

\paragraph{LiveVQA} \citep{fu2025livevqa} aggregates 3,602 image-question pairs from major news outlets, including CNN and BBC, spanning 14 thematic categories, and emphasizes temporal reasoning over recently published content.

\paragraph{MAT-Search} \citep{liu2025visual} is a manually crafted and human-verified benchmark of 150 examples designed to evaluate agentic multimodal reasoning, requiring models to handle composite multi-hop queries, retrieve external information, and invoke tools effectively across varying reasoning depths.

%% file: appendix/4_strategy_library.tex
\section{Reasoning Candidate Library}
\label{app:strategy-library}

The default library contains six manually written reasoning candidates (Table~\ref{tab:strategy-library}).
Each entry is a single short natural-language statement describing \emph{how to make progress from the current state}.
The library is fixed throughout training and evaluation.

\begin{table}[!ht]
\centering
\caption{The six-candidate multimodal-search reasoning library.}
\label{tab:strategy-library}
\small
\renewcommand{\arraystretch}{1.4}
\begin{tabular}{@{}l p{0.65\linewidth}@{}}
\toprule
\textbf{Name} & \textbf{Reasoning candidate} \\
\midrule
\textsc{general-knowledge} &
I recognize the main subject of this image and I already know the answer from general knowledge, so I can answer the question directly. \\[4pt]

\textsc{answer-from-image} &
The information needed to answer is clearly visible in the image itself, so I can determine the answer by reading it directly from what I see. \\[4pt]

\textsc{examine-image-detail} &
The part of the image that matters for this question is small or hard to make out, so I should examine that region more closely before I decide. \\[4pt]

\textsc{identify-by-image} &
I cannot confidently tell what the specific entity in this image is from its appearance alone, so I should identify it by searching with the image itself before answering. \\[4pt]

\textsc{lookup-fact} &
I can tell what the entity in the image is, but I do not know the particular fact the question is asking about, so I should look that information up. \\[4pt]

\textsc{answer-from-evidence} &
The details I have gathered so far already contain what the question is asking for, so I can now give the final answer. \\
\bottomrule
\end{tabular}
\end{table}

%% file: appendix/0_training_details.tex
\section{Training Details}
\label{app:training-details}

\paragraph{Reference recipe and our adaptation.}
We adapt the multimodal-search setup of TAPO~\citep{dong2026tapo}.
The agent is trained directly with RL, without an SFT warm-up, using VeRL, a global batch size of 64, learning rate $1\times10^{-5}$, and $8$ rollouts per GRPO group.
Each trajectory permits at most 10 interaction rounds during training, 8,192 generated tokens per round, and 32,768 tokens across the full trajectory.
If it exceeds any of these, it will be marked as incorrect.
Text retrieval during training uses a local Wikipedia retrieval index to reduce cost and evaluation uses web search through Serp API.

\paragraph{Action space.}
The agent can choose among three information-acquisition tools and a terminal answer: (i) \texttt{image\_search\_tool}, which performs reverse-image search on the question image and returns image captions; (ii) \texttt{text\_search\_tool(query)}, which performs query-specified search and returns a Qwen3-32B-generated summary; (iii) \texttt{image\_zoom\_in\_tool(bbox)}, which crops a bounding box and returns the resulting image patch; and (iv) a terminal \texttt{<answer>} action.
At every non-terminal turn, the base protocol expects a thinking block followed by exactly one \texttt{<tool\_call>} conforming to the JSON schema; SSR supplies the thinking block through candidate selection while retaining the same tool-call schema.

\paragraph{Reward.}
For RL, we use a combination of accuracy reward and format reward.
\begin{equation}
    r_\text{final} = r_\text{accuracy} + 0.5r_\text{format}
\end{equation}
Non-multiple-choice questions are judged by Qwen3-VL-32B-Instruct with greedy decoding and temperature zero given the image, question, ground truth, and the final answer generated by the model. 
Multiple-choice questions use exact match.
Format reward is model-dependent: all conditions have to generate valid \texttt{<think>} delimiters and correct tool call format; SSR models (including the index-only variant) additionally have to use the expected tool of each reasoning that they select for full format reward, otherwise they only get partial format reward for the correct tool call format; the autoregressive structured reasoning model gets the format reward only if it generates reasoning in the library with exact match.

%% file: appendix/3_reasoning_comparison.tex
\section{Reasoning Comparison: No Thinking, Freeform, and SSR}
\label{app:grpo-reasoning-comparison}

We compare no thinking, freeform reasoning, and SSR (ours) under GRPO with the Qwen3-VL-4B backbone.

\begin{figure}[!ht]
  \centering
  \includegraphics[width=\textwidth]{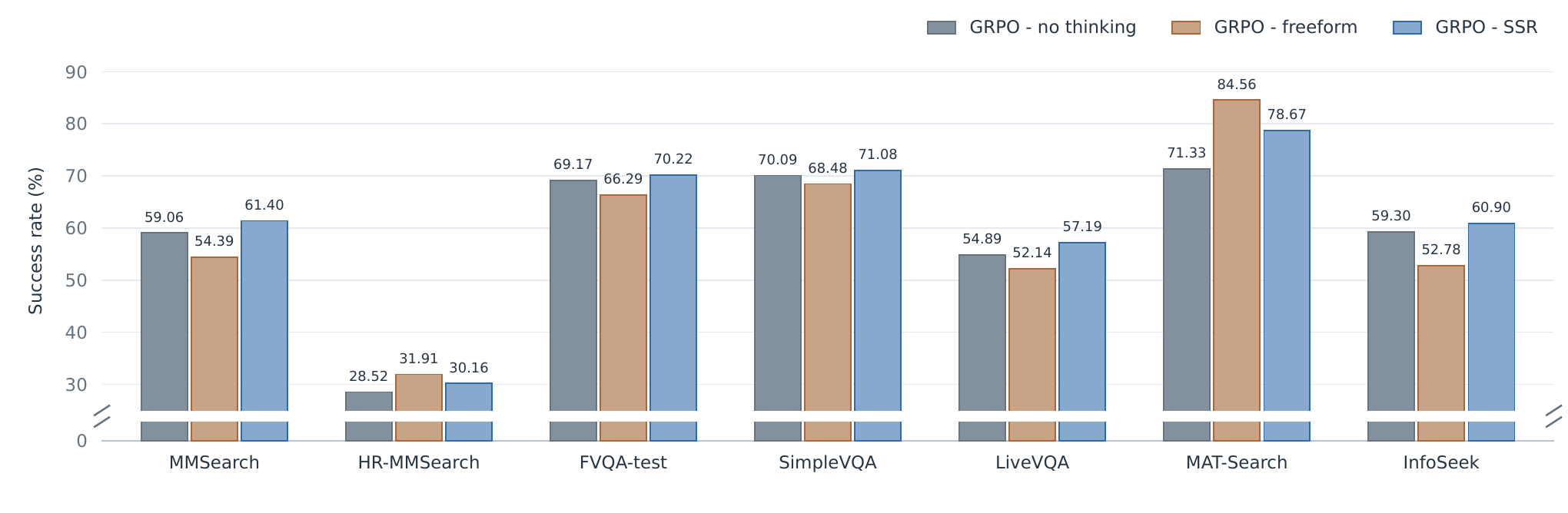}
  \caption{Success rate (\%) of the three reasoning types on the multimodal search benchmarks.}
  \label{fig:grpo-reasoning-comparison}
\end{figure}

\paragraph{Aggregate comparison.}
Figure~\ref{fig:grpo-reasoning-comparison} compares no thinking, freeform reasoning, and SSR under GRPO.
The unweighted mean success rate across the seven benchmarks is $58.91\%$, $58.65\%$, and $61.37\%$, respectively.
SSR therefore improves the benchmark mean by $2.47$ percentage points over no thinking and $2.72$ points over freeform reasoning.
Freeform reasoning is $0.26$ points below no thinking on this aggregate, so generating a freeform reasoning trace does not by itself consistently improve task performance in this comparison.

\paragraph{Consistent gains over no thinking.}
SSR outperforms no thinking on all seven benchmarks (7/7), with gains ranging from $0.99$ to $7.34$ percentage points.
In contrast, freeform reasoning improves over no thinking on only two of seven benchmarks with the same Qwen3-VL-4B backbone under GRPO.
SSR improves over no thinking across all seven benchmarks whereas freeform does not outperform no thinking on all of the benchmarks.
SSR also exceeds freeform reasoning on five of seven benchmarks, with HR-MMSearch and MAT-Search favoring freeform reasoning.

%% file: references.bib
@inproceedings{arora2026training,
  title={Training language models to reason efficiently},
  author={Arora, Daman and Zanette, Andrea},
  volume={38},
  pages={60770--60808},
  year={2025},
  booktitle={Advances in Neural Information Processing Systems}
}

@inproceedings{aytes2025sketch,
  title={Sketch-of-thought: Efficient llm reasoning with adaptive cognitive-inspired sketching},
  author={Aytes, Simon A and Baek, Jinheon and Hwang, Sung Ju},
  booktitle={Proceedings of the 2025 Conference on Empirical Methods in Natural Language Processing},
  pages={24296--24320},
  year={2025}
}

@article{bai2025qwen25vl,
  title={Qwen2.5-VL Technical Report},
  author={Bai, Shuai and Chen, Keqin and Liu, Xuejing and Wang, Jialin and Ge, Wenbin and Song, Sibo and Dang, Kai and Wang, Peng and Wang, Shijie and Tang, Jun and others},
  journal={arXiv preprint arXiv:2502.13923},
  year={2025}
}

@inproceedings{chen2023infoseek,
  title={Can pre-trained vision and language models answer visual information-seeking questions?},
  author={Chen, Y. and Hu, H. and Luan, Y. and Sun, H. and Changpinyo, S. and Ritter, A. and Chang, M.},
  booktitle={Proceedings of the 2023 Conference on Empirical Methods in Natural Language Processing},
  pages={14948--14968},
  year={2023}
}

@inproceedings{cheng2025simplevqa,
  title={SimpleVQA: Multimodal factuality evaluation for multimodal large language models},
  author={Cheng, X. and Zhang, W. and Zhang, S. and Yang, J. and Guan, X. and Wu, X. and Li, X. and Zhang, G. and Liu, J. and Mai, Y. and others},
  booktitle={Proceedings of the IEEE/CVF International Conference on Computer Vision},
  pages={4637--4646},
  year={2025}
}

@article{chng2025sensenova,
  title={SenseNova-MARS: Empowering multimodal agentic reasoning and search via reinforcement learning},
  author={Chng, Y. X. and Hu, T. and Tong, W. and Li, X. and Chen, J. and Yu, H. and Lu, J. and Guo, H. and Deng, H. and Xie, C. and others},
  journal={arXiv preprint arXiv:2512.24330},
  year={2025}
}

@inproceedings{dong2025xgrammar,
  title={{XGrammar}: Flexible and Efficient Structured Generation Engine for Large Language Models},
  author={Dong, Yixin and Ruan, Charlie F. and Cai, Yaxing and Lai, Ruihang and Xu, Ziyi and Zhao, Yilong and Chen, Tianqi},
  booktitle={Proceedings of Machine Learning and Systems},
  year={2025},
  url={https://arxiv.org/abs/2411.15100}
}

@article{dong2026tapo,
  title={TAPO: Tool-Aware Policy Optimization via Credit Transfer for Multimodal Search Agents},
  author={Dong, Chengqi and Yue, Chuhuai and He, Hang and Liu, Yandong and Tang, Fenghe and Zhou, S Kevin and Wang, Xiaohan and Chai, Jiajun and Yin, Guojun},
  journal={arXiv preprint arXiv:2606.05784},
  year={2026}
}

@article{fang2026budgetsearch,
  title={Inference-Time Budget Control for {LLM} Search Agents},
  author={Fang, Zhengru and Hu, Senkang Forest and Chang, Zhonghao and Guo, Yu and Tao, Yihang and Liu, Hongyao and Ruan, Mengzhe and Huang, Jun and Fang, Yuguang},
  journal={arXiv preprint arXiv:2605.05701},
  year={2026}
}

@inproceedings{fu2024autoguide,
  title={{AutoGuide}: Automated Generation and Selection of Context-Aware Guidelines for Large Language Model Agents},
  author={Fu, Yao and Kim, Dong-Ki and Kim, Jaekyeom and Sohn, Sungryull and Logeswaran, Lajanugen and Bae, Kyunghoon and Lee, Honglak},
  booktitle={Advances in Neural Information Processing Systems},
  volume={37},
  year={2024},
  url={https://papers.nips.cc/paper/2024/hash/d8efbb5dd415974eb095c3f06bff1f48-Abstract-Conference.html}
}

@article{fu2025livevqa,
  title={LiveVQA: Live visual knowledge seeking},
  author={Fu, Mingyang and Peng, Yuyang and Liu, Benlin and Wan, Yao and Chen, Dongping},
  journal={arXiv preprint arXiv:2504.05288v1},
  year={2025},
  note={Version 1},
  url={https://arxiv.org/abs/2504.05288v1}
}

@article{gao2025sapo,
  title={Soft Adaptive Policy Optimization},
  author={Gao, Chang and Zheng, Chujie and Chen, Xiong-Hui and Dang, Kai and Liu, Shixuan and Yu, Bowen and Yang, An and Bai, Shuai and Zhou, Jingren and Lin, Junyang},
  journal={arXiv preprint arXiv:2511.20347},
  year={2025},
  url={https://arxiv.org/abs/2511.20347}
}

@inproceedings{geng2023grammar,
  title={Grammar-constrained decoding for structured NLP tasks without finetuning},
  author={Geng, Saibo and Josifoski, Martin and Peyrard, Maxime and West, Robert},
  booktitle={Proceedings of the 2023 Conference on Empirical Methods in Natural Language Processing},
  pages={10932--10952},
  year={2023}
}

@article{geng2025webwatcher,
  title={WebWatcher: Breaking New Frontier of Vision-Language Deep Research Agent},
  author={Geng, Xinyu and Xia, Peng and Zhang, Zhen and Wang, Xinyu and Wang, Qiuchen and Ding, Ruixue and Wang, Chenxi and Wu, Jialong and Zhao, Yida and Li, Kuan and Jiang, Yong and Xie, Pengjun and Huang, Fei and Zhou, Jingren},
  journal={arXiv preprint arXiv:2508.05748},
  year={2025}
}

@misc{google2025gemini3flash,
  title={{Gemini 3 Flash} Model Card},
  author={{Google DeepMind}},
  howpublished={\url{https://deepmind.google/models/model-cards/gemini-3-flash/}},
  month={December},
  year={2025}
}

@misc{google2026gemini31pro,
  title={{Gemini 3.1 Pro} Model Card},
  author={{Google DeepMind}},
  howpublished={\url{https://deepmind.google/models/model-cards/gemini-3-1-pro/}},
  month={February},
  year={2026}
}

@misc{google2026gemini3pro,
  title={{Gemini 3 Pro} Model Card},
  author={{Google DeepMind}},
  howpublished={\url{https://deepmind.google/models/model-cards/gemini-3-pro/}},
  month={May},
  year={2026}
}

@inproceedings{han2025tokenbudget,
  title={Token-Budget-Aware {LLM} Reasoning},
  author={Han, Tingxu and Wang, Zhenting and Fang, Chunrong and Zhao, Shiyu and Ma, Shiqing and Chen, Zhenyu},
  year={2025},
  booktitle={Findings of the Association for Computational Linguistics: ACL 2025},
  pages={24842--24855},
  doi={10.18653/v1/2025.findings-acl.1274},
  url={https://aclanthology.org/2025.findings-acl.1274/}
}

@inproceedings{he2024webvoyager,
  title={WebVoyager: Building an end-to-end web agent with large multimodal models},
  author={He, Hongliang and Yao, Wenlin and Ma, Kaixin and Yu, Wenhao and Dai, Yong and Zhang, Hongming and Lan, Zhenzhong and Yu, Dong},
  booktitle={Proceedings of the 62nd Annual Meeting of the Association for Computational Linguistics (Volume 1: Long Papers)},
  year={2024}
}

@inproceedings{hong2025deepeyesv2,
  title={{DeepEyesV2}: Toward Agentic Multimodal Model},
  author={Hong, Jack and Zhao, Chenxiao and Zhu, ChengLin and Lu, Weiheng and Xu, Guohai and Yu, Xing},
  booktitle={International Conference on Learning Representations},
  year={2026}
}

@article{jiang2024mmsearch,
  title={MMSearch: Benchmarking the potential of large models as multi-modal search engines},
  author={Jiang, D. and Zhang, R. and Guo, Z. and Wu, Y. and Lei, J. and Qiu, P. and Lu, P. and Chen, Z. and Fu, C. and Song, G. and others},
  journal={arXiv preprint arXiv:2409.12959},
  year={2024}
}

@inproceedings{lai2026mini,
  title={Mini-o3: Scaling up reasoning patterns and interaction turns for visual search},
  author={Lai, Xin and Li, Junyi and Li, Wei and Liu, Tao and Li, Tianjian and Zhao, Hengshuang},
  booktitle={International Conference on Learning Representations},
  volume={2026},
  pages={76722--76746},
  year={2026}
}

@article{li2026hypereyes,
  title={HyperEyes: Dual-Grained Efficiency-Aware Reinforcement Learning for Parallel Multimodal Search Agents},
  author={Li, Guankai and Chen, Jiabin and Xu, Yi and Zhang, Xichen and Lu, Yuan},
  journal={arXiv preprint arXiv:2605.07177},
  year={2026}
}

@article{liu2025visual,
  title={Visual agentic reinforcement fine-tuning},
  author={Liu, Z. and Zang, Y. and Zou, Y. and Liang, Z. and Dong, X. and Cao, Y. and Duan, H. and Lin, D. and Wang, J.},
  journal={arXiv preprint arXiv:2505.14246},
  year={2025}
}

@article{ma2025nothinking,
  title={Reasoning Models Can Be Effective Without Thinking},
  author={Ma, Wenjie and He, Jingxuan and Snell, Charlie and Griggs, Tyler and Min, Sewon and Zaharia, Matei},
  journal={arXiv preprint arXiv:2504.09858},
  year={2025}
}

@article{narayan2025deepmmsearch,
  title={DeepMMSearch-R1: Empowering multimodal LLMs in multimodal web search},
  author={Narayan, K. and Xu, Y. and Cao, T. and Nerella, K. and Patel, V. M. and Shiee, N. and Grasch, P. and Jia, C. and Yang, Y. and Gan, Z.},
  journal={arXiv preprint arXiv:2510.12801},
  year={2025}
}

@misc{openai2024gpt4o,
  title={Hello {GPT-4o}},
  author={{OpenAI}},
  howpublished={\url{https://openai.com/index/hello-gpt-4o/}},
  month={May},
  year={2024}
}

@misc{openai2024gpt4omini,
  title={{GPT-4o mini}: Advancing Cost-Efficient Intelligence},
  author={{OpenAI}},
  howpublished={\url{https://openai.com/index/gpt-4o-mini-advancing-cost-efficient-intelligence/}},
  month={July},
  year={2024}
}

@misc{openai2025gpt5,
  title={Introducing {GPT-5}},
  author={{OpenAI}},
  howpublished={\url{https://openai.com/index/introducing-gpt-5/}},
  month={August},
  year={2025}
}

@misc{openai2025gpt52,
  title={Introducing {GPT-5.2}},
  author={{OpenAI}},
  url={https://openai.com/index/introducing-gpt-5-2/},
  month={December},
  year={2025}
}

@inproceedings{ouyang2026reasoningbank,
  title={Reasoningbank: Scaling agent self-evolving with reasoning memory},
  author={Ouyang, Siru and Yan, Jun and Hsu, I and Chen, Yanfei and Jiang, Ke and Wang, Zifeng and Han, Rujun and Le, Long and Daruki, Samira and Tang, Xiangru and others},
  booktitle={International Conference on Learning Representations},
  volume={2026},
  pages={94327--94354},
  year={2026}
}

@article{shao2024deepseekmath,
  title={DeepSeekMath: Pushing the limits of mathematical reasoning in open language models},
  author={Shao, Zhihong and Wang, Peiyi and Zhu, Qihao and Xu, Runxin and Song, Junxiao and Bi, Xiao and Zhang, Haowei and Zhang, Mingchuan and Li, Y. K. and Wu, Y. and Guo, Daya},
  journal={arXiv preprint arXiv:2402.03300},
  year={2024}
}

@article{sui2025stop,
  title={Stop Overthinking: A Survey on Efficient Reasoning for Large Language Models},
  author={Sui, Yang and Chuang, Yu-Neng and Wang, Guanchu and Zhang, Jiamu and Zhang, Tianyi and Yuan, Jiayi and Liu, Hongyi and Wen, Andrew and Zhong, Shaochen and Zou, Na and Chen, Hanjie and Hu, Xia},
  journal={arXiv preprint arXiv:2503.16419},
  year={2025}
}

@article{sun2026llm,
  title={LLM Agents Already Know When to Call Tools--Even Without Reasoning},
  author={Sun, Chung-En and Liu, Linbo and Yan, Ge and Wang, Zimo and Weng, Tsui-Wei},
  journal={arXiv preprint arXiv:2605.09252},
  year={2026}
}

@misc{team2025qwen3vl,
  title={{Qwen3-VL}: Sharper Vision, Deeper Thought, Broader Action},
  author={{Qwen Team}},
  howpublished={\url{https://qwen.ai/blog?from=research.latest-advancements-list&id=99f0335c4ad9ff6153e517418d48535ab6d8afef}},
  year={2025}
}

@article{wang2023voyager,
  title={Voyager: An open-ended embodied agent with large language models},
  author={Wang, Guanzhi and Xie, Yuqi and Jiang, Yunfan and Mandlekar, Ajay and Xiao, Chaowei and Zhu, Yuke and Fan, Linxi and Anandkumar, Anima},
  journal={arXiv preprint arXiv:2305.16291},
  year={2023}
}

@inproceedings{wang2025workflow,
  title={Agent Workflow Memory},
  author={Wang, Zora Zhiruo and Mao, Jiayuan and Fried, Daniel and Neubig, Graham},
  booktitle={Proceedings of the 42nd International Conference on Machine Learning},
  volume={267},
  pages={63897--63911},
  year={2025},
  url={https://proceedings.mlr.press/v267/wang25bx.html}
}

@article{wu2025mmsearch,
  title={MMSearch-R1: Incentivizing LMMs to search},
  author={Wu, J. and Deng, Z. and Li, W. and Liu, Y. and You, B. and Li, B. and Ma, Z. and Liu, Z.},
  journal={arXiv preprint arXiv:2506.20670},
  year={2025}
}

@article{xu2025chain,
  title={{Chain of Draft}: Thinking Faster by Writing Less},
  author={Xu, Silei and Xie, Wenhao and Zhao, Lingxiao and He, Pengcheng},
  journal={arXiv preprint arXiv:2502.18600},
  year={2025}
}

@article{xue2026strata,
  title={{StraTA}: Incentivizing Agentic Reinforcement Learning with Strategic Trajectory Abstraction},
  author={Xue, Xiangyuan and Zhou, Yifan and Wang, Zidong and Tang, Shengji and Torr, Philip and Ouyang, Wanli and Bai, Lei and Yin, Zhenfei},
  journal={arXiv preprint arXiv:2605.06642},
  year={2026}
}

@article{yan2026prommsearch,
  title={{ProMMSearchAgent}: A Generalizable Multimodal Search Agent Trained with Process-Oriented Rewards},
  author={Yan, Wentao and Wang, Shengqin and Zhou, Huichi and Chen, Yihang and Shao, Kun and Xie, Yuan and Zhang, Zhizhong},
  journal={arXiv preprint arXiv:2604.20486},
  year={2026}
}

@article{yang2024buffer,
  title={Buffer of Thoughts: Thought-augmented reasoning with large language models},
  author={Yang, Ling and Yu, Zhaochen and Zhang, Tianjun and Cao, Shiyi and Xu, Minkai and Zhang, Wentao and Gonzalez, Joseph E. and Cui, Bin},
  journal={arXiv preprint arXiv:2406.04271},
  year={2024}
}

@article{yang2026skillopt,
  title={SkillOpt: Executive Strategy for Self-Evolving Agent Skills},
  author={Yang, Yifan and Gong, Ziyang and Huang, Weiquan and Yang, Qihao and Zhou, Ziwei and Huang, Zisu and Li, Yan and Gao, Xuemei and Dai, Qi and Liu, Bei and Qiu, Kai and Yang, Yuqing and Chen, Dongdong and Yang, Xue and Luo, Chong},
  journal={arXiv preprint arXiv:2605.23904},
  year={2026}
}

@article{yao2022react,
  title={React: Synergizing reasoning and acting in language models},
  author={Yao, Shunyu and Zhao, Jeffrey and Yu, Dian and Du, Nan and Shafran, Izhak and Narasimhan, Karthik and Cao, Yuan},
  journal={arXiv preprint arXiv:2210.03629},
  year={2022}
}

@article{zhang2025skywork,
  title={Skywork-R1V4: Toward agentic multimodal intelligence through interleaved thinking with images and deepresearch},
  author={Zhang, Y. and Hu, L. and Sun, H. and Wang, P. and Wei, Y. and Yin, S. and Pei, J. and Shen, W. and Xia, P. and Peng, Y. and Xie, T. and Li, E. and Liu, Y. and Song, X. and Zhou, Y.},
  journal={arXiv preprint arXiv:2512.02395},
  year={2025}
}

@article{zhang2026vsearcher,
  title={{VSearcher}: Long-Horizon Multimodal Search Agent via Reinforcement Learning},
  author={Zhang, Ruiyang and Sun, Qianguo and Song, Chao and Qi, Yiyan and Zheng, Zhedong},
  journal={arXiv preprint arXiv:2603.02795},
  year={2026}
}

@inproceedings{zheng2024sglang,
  title={{SGLang}: Efficient Execution of Structured Language Model Programs},
  author={Zheng, Lianmin and Yin, Liangsheng and Xie, Zhiqiang and Sun, Chuyue and Huang, Jeff and Yu, Cody Hao and Cao, Shiyi and Kozyrakis, Christos and Stoica, Ion and Gonzalez, Joseph E. and Barrett, Clark and Sheng, Ying},
  booktitle={Advances in Neural Information Processing Systems},
  volume={37},
  year={2024}
}

@article{zheng2025gspo,
  title={Group Sequence Policy Optimization},
  author={Zheng, Chujie and Liu, Shixuan and Li, Mingze and Chen, Xiong-Hui and Yu, Bowen and Gao, Chang and Dang, Kai and Liu, Yuqiong and Men, Rui and Yang, An and Zhou, Jingren and Lin, Junyang},
  journal={arXiv preprint arXiv:2507.18071},
  year={2025},
  url={https://arxiv.org/abs/2507.18071}
}

@article{zheng2025skillweaver,
  title={SkillWeaver: Web Agents can Self-Improve by Discovering and Honing Skills},
  author={Zheng, Boyuan and Fatemi, Michael Y. and Jin, Xiaolong and Wang, Zora Zhiruo and Gandhi, Apurva and Song, Yueqi and Gu, Yu and Srinivasa, Jayanth and Liu, Gaowen and Neubig, Graham and Su, Yu},
  journal={arXiv preprint arXiv:2504.07079},
  year={2025}
}

@inproceedings{zheng2026deepeyes,
  title={{DeepEyes}: Incentivizing ``Thinking with Images'' via Reinforcement Learning},
  author={Zheng, Ziwei and Yang, Michael and Hong, Jack and Zhao, Chenxiao and Xu, Guohai and Yang, Le and Shen, Chao and Yu, Xing},
  booktitle={International Conference on Learning Representations},
  year={2026}
}
